\documentclass[nohyperref]{article}
\usepackage[dvipsnames]{xcolor}

\usepackage[accepted]{icml2022}

\usepackage{amsmath,amsfonts,bm}

\def\eqref#1{equation~\ref{#1}}

\def\1{\bm{1}}

\DeclareMathAlphabet{\mathsfit}{\encodingdefault}{\sfdefault}{m}{sl}
\SetMathAlphabet{\mathsfit}{bold}{\encodingdefault}{\sfdefault}{bx}{n}

\DeclareMathOperator*{\argmin}{arg\,min}

\usepackage{hyperref}
\usepackage{url}
\usepackage{tikz}
\newif\ifdrafting 
\draftingtrue       
\ifdrafting
\newcommand{\AP}[1]{{\textbf{\color{purple}Alizee: #1}}}
\newcommand{\HY}[1]{{\textbf{\color{RoyalBlue}Hugo: #1}}}
\newcommand{\RK}[1]{{\textbf{\color{red}Rita: #1}}}

\newcommand{\TODO}[1]{{\textbf{\color{green} TODO: #1}}}

\else
\newcommand{\AP}[1]{}
\newcommand{\HY}[1]{}
\newcommand{\RK}[1]{}
\newcommand{\TODO}[1]{}

\fi

\usepackage[utf8]{inputenc} \usepackage[T1]{fontenc}    \usepackage{hyperref}       \usepackage{url}            \usepackage{booktabs}       \usepackage{amsfonts}       \usepackage{nicefrac}       \usepackage{microtype}      \usepackage{amsmath}
\usepackage{multirow}
\usepackage{graphicx}
\usepackage{subcaption}
\usepackage{comment}
\usepackage{booktabs}
\usepackage{graphicx}
\usepackage{amsthm}
\usepackage{stmaryrd}
\usepackage{wrapfig}
\usepackage{dsfont}
\usepackage{pifont}\newcommand{\cmark}{\ding{51}}\newcommand{\xmark}{\textcolor{lightgray}{\ding{55}}}\DeclareMathOperator{\X}{\mathbf{X}}
\newtheorem{theorem}{Proposition}
\setcitestyle{numbers,open={[},close={]}}
\usepackage{listings}

\definecolor{dkgreen}{rgb}{0,0.6,0}
\definecolor{gray}{rgb}{0.5,0.5,0.5}
\definecolor{mauve}{rgb}{0.58,0,0.82}

\icmltitlerunning{Temporal Label Smoothing }
\begin{document}

\twocolumn[
\icmltitle{Temporal Label Smoothing for Early Event Prediction}

\icmlsetsymbol{equal}{*}
\icmlsetsymbol{co-sup}{+}

\begin{icmlauthorlist}
\icmlauthor{Hugo Yèche}{equal,eth}
\icmlauthor{Alizée Pace}{equal,eth,aic,mpi}
\icmlauthor{Gunnar Rätsch}{co-sup,eth}
\icmlauthor{Rita Kuznetsova}{co-sup,eth}
\end{icmlauthorlist}

\icmlaffiliation{eth}{Department of Computer Science, ETH Zürich, Switzerland}
\icmlaffiliation{mpi}{Max Planck Institute for Intelligent Systems, Tübingen, Germany}
\icmlaffiliation{aic}{ETH AI Center, ETH Zürich,Switzerland}

\icmlcorrespondingauthor{Hugo Yèche}{hyeche@ethz.ch}

\icmlkeywords{Machine Learning, ICML}

\vskip 0.3in
]

\printAffiliationsAndNotice{\icmlEqualContribution}
\begin{abstract}
Models that can predict the occurrence of events ahead of time with low false-alarm rates are critical to the acceptance of decision support systems in the medical community. This challenging task is typically treated as a simple binary classification, ignoring temporal dependencies between samples, whereas we propose to exploit this structure. We first introduce a common theoretical framework unifying dynamic survival analysis and early event prediction. Following an analysis of objectives from both fields, we propose Temporal Label Smoothing (TLS), a simpler, yet best-performing method that preserves prediction monotonicity over time. By focusing the objective on areas with a stronger predictive signal, TLS improves performance over all baselines on two large-scale benchmark tasks. Gains are particularly notable along clinically relevant measures, such as event recall at low false-alarm rates. TLS reduces the number of missed events by up to a factor of two over previously used approaches in early event prediction.

\end{abstract}

\section{Introduction}

Early event prediction (EEP) is a time-series task concerned with determining whether an event will occur within a fixed time horizon. Key to safety-critical operations such as environmental monitoring \citep{DiGiuseppe2016}, EEP is also highly relevant to clinical decision-making, where the deployment of in-patient risk stratification models can significantly improve patient outcomes and facilitate resource planning \citep{sutton2020}. For instance, the National Early Warning Score (NEWS), a simple rule-based model predicting acute deterioration in critical care units, has been demonstrated to reduce in-patient mortality \citep{smith2013, pullyblank2020implementation}. Deteriorating patient signals are often identified by mining large quantities of existing medical data and associated patient outcomes, which has sparked a growing interest in machine learning and medical literature. Applications of such adverse event prediction models include alarm systems for delirium~\citep{wong2018development}, septic shock~\citep{fagerstrom2019lisep}, as well as circulatory or kidney failure in the intensive care unit (ICU)~\citep{hyland2020, tomavsev2019}.

Still, prediction systems often suffer from high false-alarm rates with limited usefulness in a practical context \citep{sutton2020}, despite the development of deep learning architectures addressing issues of high dimensionality, irregular sampling, or informative missingness in time-series~\citep{fagerstrom2019lisep,tomavsev2019,DBLP:conf/icml/HornMBRB20,DBLP:conf/iclr/ShuklaM21}. The typically rare occurrence and noisy definition of events of interest induce challenging, highly imbalanced datasets for model training~\citep{tomavsev2019}, yet early event prediction remains largely considered as a simple binary classification task~\citep{hyland2020, Lauritsen2020, DBLP:conf/icml/HornMBRB20, roy2021multitask}. 

In this work, we systematically study different choices of objective functions for this task and outline a novel, simple, yet best-performing approach to early event prediction. In particular, we argue that leveraging the temporal structure of early event prediction is critical to improving model performance. The dynamic survival analysis (DSA) framework~\citep{van2007dynamic}, for instance, which aims to regress the time until a unique event of interest occurs, enforces structural properties across timepoints and studied horizons~\cite{lee2019dynamic,jarrett2019dynamic}. Inspired by this, we propose to induce monotonicity in model predictions over time with Temporal Label Smoothing (TLS). This novel regularization strategy also mirrors our expected confidence in the strength of prediction signals over time.

\begin{figure}[th] \centering
  \includegraphics[width=0.9\linewidth]{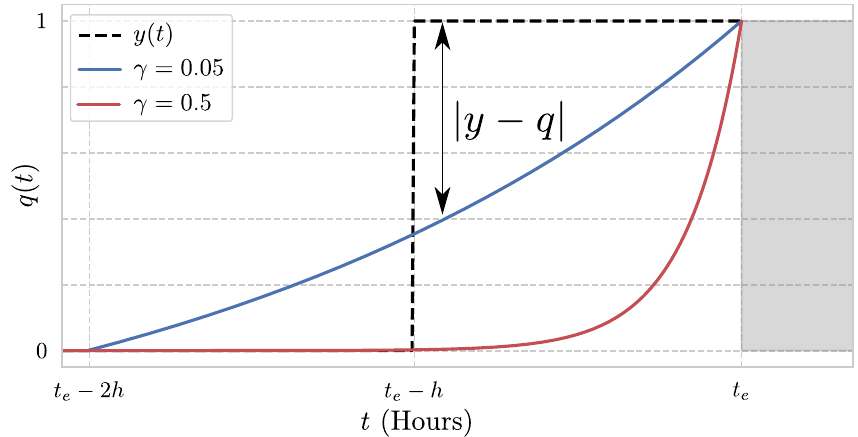}      
\caption{\textbf{Illustration of temporal label smoothing} for early event prediction. Predictions are carried out over a horizon $h$ and $t_e$ is the time of the next event, shaded in grey. True labels in black. $\gamma$ controls the smoothing strength of surrogate labels $q$.}
  \label{fig:gamma_TLS}
  \vspace{-1em}
\end{figure}

\vspace{-1em}

\paragraph{Contributions.} The contributions of our work are threefold: 
(i) First, we adapt and benchmark existing approaches from the survival literature to early event prediction, highlighting theoretical similarities between these frameworks. We bridge the gap with prior work \citep{tomavsev2019,van2007dynamic, parast2014landmark}, showing that these enforce temporal structure properties in model predictions. (i) Next, we introduce a simple method to achieve this for our single-horizon prediction framework\footnote[1]{All code is made publicly available at \url{https://anonymous.4open.science/r/tls/}.}.
(iii) Finally, we explore real-world event prediction tasks and demonstrate the performance gains of our method, particularly on clinically relevant metrics. Ablations show that this effectively focuses training on datapoints with a stronger predictive signal.

\begin{table*}[t]
    \centering
    \caption{\textbf{Related work.} Comparison of different relevant training objectives. Early event labels and model predictions at time $t$ are denoted $y_t^h = \mathds{1}[t > t_e - h]$ and $\hat{y}_t^h \in [0,1]$, dropping horizon $h$ when fixed. Hazard function labels and predictions are denoted $\lambda^h_t=\mathds{1}[t = t_e - h]$ and $\hat{\lambda}^h_t$. Temporal structure properties are time monotonicity (Eq. \ref{eq:time_monotonous}), horizon monotonicity (Eq. \ref{eq:horizon_monotonous}), and consistency (Eq. \ref{eq:time_consistency}). Additional details are provided in Appendix \ref{appendix:relatedwork}.} \label{tab:related_work}
\begin{tabular}{lccccc}
    \toprule
        \multirow{2}{*}{Training objective} & Previously used  & \multicolumn{3}{c}{Temporal structure} &Loss function, \\  & for event prediction & \hspace*{1mm}(\ref{eq:time_monotonous}) \hspace*{1mm} & \hspace*{1mm}(\ref{eq:horizon_monotonous}) \hspace*{1mm}& \hspace*{1mm}(\ref{eq:time_consistency}) \hspace*{1mm}& summed over label values\\ \midrule 
         Cross-entropy \citep{Lauritsen2020, hyland2020} & \cmark & \xmark & \xmark& \xmark& $ - \sum_t  y_t \log (\hat{y}_t) $\\ Balanced cross-entropy \cite{king2001logistic} & \cmark & \xmark & \xmark& \xmark& $- \sum_t \omega y_t \log (\hat{y}_t) $\\ 

        Focal loss \cite{lin2017,wang2020feature, roy2022disability}  & \cmark & \xmark& \xmark&\xmark & $ - \sum_t\omega (1-\hat{y}_t)^{\zeta} y_t \log (\hat{y}_t) $\\ 

        Label smoothing \cite{DBLP:conf/cvpr/SzegedyVISW16}  & \xmark & \xmark & \xmark& \xmark& $ - \sum_t q^{LS} (y_t) \log (\hat{y}_t) $\\ 

        Multi-horizon prediction \cite{tomavsev2019, jarrett2019dynamic} & \cmark & \xmark& \cmark& \xmark &  $ - \sum_t \sum_h y_t^h \log(\hat{y}_t^h) $\\ \midrule
        Survival analysis likelihood \citep{cox1972,kalbfleisch2011statistical}  & \xmark & \xmark & \cmark& \xmark& $ - \sum_h \lambda_0^h \log(\hat{\lambda}_0^h)$ \\
        
        Landmarking \citep{van2007dynamic,parast2014landmark} & \xmark&  \xmark  & \cmark & \cmark &  $ - \sum_t \sum_h \lambda_t^h \log(\hat{\lambda}_t^h) $ \\ 

        TCSR \citep{Maystre2022}  & \xmark& \xmark&\cmark & \cmark   & $ - \sum_t \sum_h \hat{\lambda}_{t+1}^{h -1} \log(\hat{\lambda}_t^h)$ \\

        \midrule
        \textbf{Temporal label smoothing}  &  \cmark &  \cmark& \xmark & \xmark & $ - \sum_t q^{TLS}(t)\log (\hat{y}_t) $ \\ \bottomrule
    \end{tabular}
\end{table*}

\section{Problem formalism and related work}\label{sec:pf_relatedwork}

We start by formalizing the early event prediction task and highlight its similarities and distinctions with survival analysis. After discussing its typical training objectives, we outline some temporal structure properties induced by label definition -- which lead to novel optimization objectives.

\subsection{Early event prediction (EEP)}\label{sec:problem_formalism}

We assume access to a dataset of irregular time series of covariates $\mathbf{x}_{i,t}$ and binary event labels $e_{i,t}$ encoding whether an event of interest is occurring at time $t$ in series of index $i$. Each sample is a sequence $ \{ (\mathbf{x}_{i,1}, e_{i,1}), \ldots, (\mathbf{x}_{i,T_i}, e_{i,T_i}) \}$ of length $T_i$. In the clinical setting, this could correspond to individual patient trajectories as time series of observations, with labeled events such as organ failure or death. For clarity, we drop index $i$ unless explicitly needed.

For each point $t$ along a time series, the covariates observed up to this point are denoted $\X_{t} = [\mathbf{x}_{0}, \ldots,\mathbf{x}_{t}]$ and the absolute time of the next event is given by $t_e = \argmin_{\tau: \tau \geq t} \{e_{\tau}: e_{\tau} = 1 \}$. Our task consists of modeling the probability of this event occurring within a fixed prediction horizon $h$: $y^h(t) = P(t>t_e - h |\mathbf{X}_t)$. In practice, we only access hard, binary labels $y^h(t) = \mathds{1}\left[t>t_e - h \right]$. Estimates of this event probability, denoted $\hat{y}^h(t)$, are typically obtained by maximizing label likelihood through binary classification. As our task focuses specifically on early modeling, no prediction is carried out if the event is currently occurring.

\paragraph{Comparison to survival analysis.} Both early event prediction and survival analysis are concerned with modeling the occurrence of an event of interest. These tasks differ in their variable of interest when applied to time series. Survival analysis is focused on studying event probability as a function of time-to-event $h$ for a fixed timepoint $t$. It aims at modeling the survival function $S(h \vert \mathbf{X}_t) = P(t_e-t > h \vert \mathbf{X}_t)$. Early event prediction, in contrast, is concerned with event probability as a function of time $t$ for a fixed horizon $h$. As a result, for a fixed $\{t,h\}$ and under the assumption of an event occurring only once, we have: $y^h(t) = 1- S(h \vert \mathbf{X}_t) $. A dynamic survival analysis (DSA) model could therefore be used for EEP, fixing the horizon to that of interest.  This leads to a first experimental question: can the survival objective, which considers all event horizons, improve performance on early event prediction at fixed $h$?

\subsection{Optimization objectives for EEP} 

We compare relevant training objectives for early event prediction in Table \ref{tab:related_work}, with further detail in Appendix~\ref{appendix:relatedwork}. Prior work on EEP typically focuses on addressing issues of class imbalance through loss reweighting techniques. Static class reweighting was used for sepsis or circulatory failure prediction~\citep{futoma2017learning,hyland2020} through a balanced cross-entropy, which assigns a higher weight to samples from the minority class~\citep{king2001logistic}. Still, performance improvements with this objective remain limited on highly imbalanced prediction tasks~\citep{yeche2021}. In contrast, dynamic reweighting methods such as focal loss and extensions~\citep{lin2017, polyloss} induce a learning bias towards samples with high model uncertainty, typically harder to classify. This approach can improve the prediction of disease progression from imbalanced datasets~\citep{wang2020feature,roy2022disability} but does not consider patterns of sample informativeness over time. Whereas class-imbalance techniques are not designed to account for any temporal structure between samples, these methods give higher importance to positive samples from the minority class, which are located closer to the event.

\subsection{Preserving temporal structure}
\label{sec:temp_structure}
In this section, we highlight how different frameworks for early event prediction or dynamic survival analysis enforce some temporal structure properties induced by the task.

\paragraph{Temporal structure.} Another important distinction must be made between early event prediction and typical classification tasks, in which data is independent and identically distributed (i.i.d.). Both in EEP and in survival analysis, labels are dependent over time. Within a patient stay, the design of our task induces the following temporal structure properties:
 \begin{align}
 \text{Time monotonicity: } &y^h(t) \leq y^h(t+\delta t) \label{eq:time_monotonous} \\ 
\text{ Horizon monotonicity: } &y^h(t) \geq y^{h+\delta h}(t) \label{eq:horizon_monotonous} \\
\text{Consistency: } & y^h (t) = y^{h - \delta t} (t+\delta t)   \label{eq:time_consistency}
\end{align}
for $\delta t, \delta h >0 $. Note that each property can be obtained from the other two.

\paragraph{Temporally structured objectives.} Some early event prediction and survival analysis objectives induce the above structural properties in model predictions.

In multi-horizon prediction (MHP), the EEP framework is modified to output event predictions over multiple horizons~\cite{tomavsev2019,tomavsev2021,roy2021multitask}. Predictions are enforced to be monotonically decreasing over the horizon~\citep{tomavsev2019}, such that if $h \leq h'$, then $\hat{y}^{h}(t) \geq \hat{y}^{h'}(t)$, as in Eq. \ref{eq:horizon_monotonous}. This has been shown to improve event prediction performance on the horizon of interest $h$.

Survival analysis also enforces horizon monotonicity if the survival function is modeled through the hazard function, defined as $\lambda(h \vert \mathbf{X}_t) = P(t_e-t = h | t_e-t \geq h, \mathbf{X}_t)$.
The survival likelihood can then be maximized through binary cross-entropy on the hazard function~\cite{kalbfleisch2011statistical,craig2021survival}, recovering survival as follows: $S(h \vert \mathbf{X}_t) =  \prod_{k=1}^h \left( 1 - \lambda(k \vert \mathbf{X}_t)\right)$. Equation~\ref{eq:horizon_monotonous} is enforced by the positivity of the hazard. Interestingly, recent works in DSA directly model the survival function~\cite{lee2019dynamic,jarrett2019dynamic}, and lose this temporal inductive bias.

Methods extending survival analysis to the dynamics setting~\cite{van2007dynamic}, where $t$ is allowed to vary, are designed to enforce temporal consistency across horizons (Eq.~\ref{eq:time_consistency} can also be written in terms of the hazard function). 
For each timestep $t$ in the training data, landmarking adjusts the prediction horizon to $h-t$, learning the hazard distribution for all horizons and timesteps jointly~\cite{van2007dynamic, parast2014landmark}. A parallel can be drawn between multi-horizon prediction in EEP and landmarking in DSA, with a key difference in the likelihood considered: MHP maximizes event prediction probability, whereas landmarking deals with hazards.

Finally, whereas landmarking induces temporal consistency across labels, \citet{Maystre2022} directly enforces consistency across hazard predictions $\hat{\lambda}(h| \mathbf{X}_t)$. This can be achieved through dynamic programming, substituting ground truth labels with predictions from following time steps.

Overall, all methods discussed enforce forms of temporal monotonicity or consistency over \textit{horizons} (Eqs. \ref{eq:horizon_monotonous} and  \ref{eq:time_consistency}). In contrast, Equation \ref{eq:time_monotonous} is most relevant to early event prediction, where $h$ is fixed: we propose a training objective explicitly designed to preserve this form of temporal structure.

\begin{figure*}[t]
    \centering
    \begin{subfigure}[b]{0.45\textwidth}
      \includegraphics[width=\textwidth]{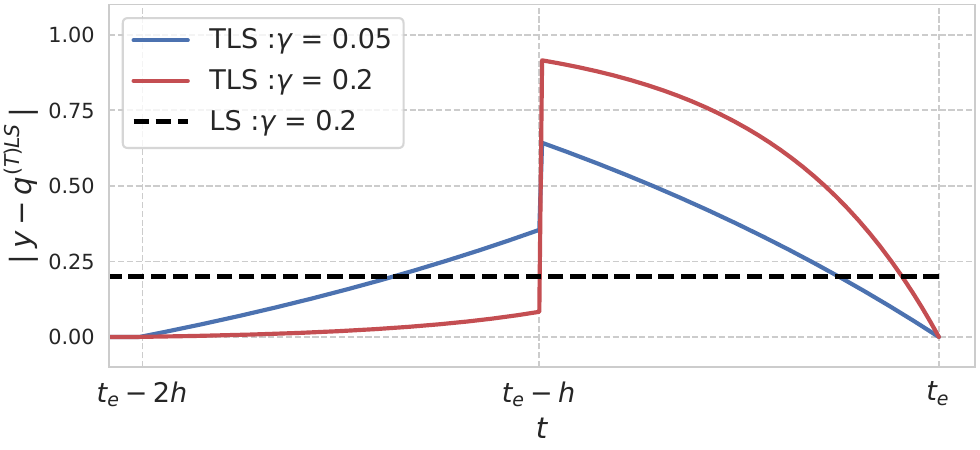}\vspace{-0.5em}
      \caption{Parametrization $q^{exp}$.}
    \end{subfigure} \hspace{1em}
    \begin{subfigure}[b]{0.45\textwidth}
      \includegraphics[width=\textwidth]{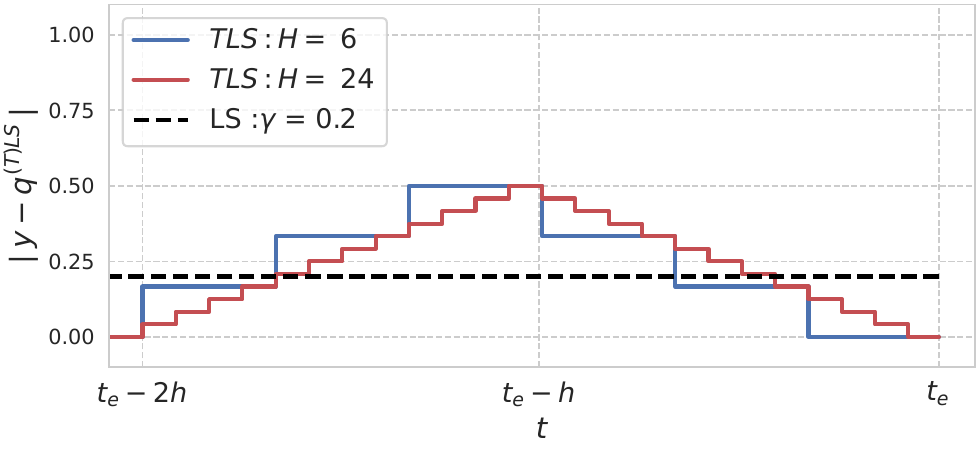}\vspace{-0.5em} \caption{Parametrization $q^{step}$.}\label{fig:smoothing_MHP} 
    \end{subfigure}
    \caption{\textbf{Label smoothing strength over time} under different parametrizations, with $(h_{min},h_{max})=(0,2h)$. Note that $|y-q^{(T)LS}|$ corresponds to the difference in optimum $y^*$ between the smoothed objective and cross-entropy. The black dashed line represents  this difference for regular label smoothing. Smoothing function $q^{step}$ is equivalent to multi-horizon prediction with a unique output.}
    \label{fig:smoothing_strength}
    \vspace{-1em}
\end{figure*}

\section{Temporal label smoothing}\label{sec:TLS} 

We introduce temporal label smoothing, our approach to enforce the structural property most relevant to our task (Eq.~\ref{eq:time_monotonous}). Thanks to prior knowledge of the labels' structure, our approach focuses training on relevant timesteps and overcomes issues with noisy label boundaries. 

Temporal label smoothing substitutes the original label distribution $y$ in the cross-entropy objective with a time-dependent distribution $q(t)$. We constrain this surrogate target to be \textit{monotonically increasing with time}. In practice, as illustrated in Figure~\ref{fig:smoothing_strength}, this increases smoothing strength around the label boundary $t_e - h $, reducing prediction certainty in this region, which we show to be prone to high error rates in Section \ref{sec:perf-results}.

Recent work in dynamic survival analysis also proposes to replace labels in the training objective \citep{Maystre2022}, this time with predictions at different time points to enforce temporal consistency (Eq. \ref{eq:time_consistency}). In practice, as demonstrated experimentally in Section \ref{sec:results}, we find this approach to be unstable and to converge poorly on real datasets with long time series and large event horizons. In contrast, we propose to replace labels with a prediction-independent distribution fixed \textit{a priori}, and thus less prone to optimization challenges. 

\paragraph{Smoothing parametrizations.} We propose various temporal smoothing parametrizations for $q_{t}$ in Appendix \ref{appendix:temporal_smoothing_fn}. Experimental results suggest that an exponential parametrization, defined as follows, performs best on considered tasks. \begin{equation*}
\resizebox{0.45\textwidth}{!}{$
    q^{exp}(t)  = \begin{cases}
    0 & \text{if }  t \leq t_e - h_{max} \\
    e^{ - \gamma ( t_e - t - d) } + A & \text{if } t_e - h_{max} < t < t_e - h_{min} \\ 
    1 & \text{if }   t \geq t_e - h_{min}
    
    \end{cases}$}
\end{equation*}
Parameters $h_{min}$ and $h_{max}$ define the time range over which we apply smoothing, namely $[t_e - h_{max}$, $t_e - h_{min}]$. Under this constraint, parameters $\{d, A\}$ are defined to enforce $q_{t}$ to be continuous at boundary points (see Appendix \ref{appendix:temporal_smoothing_fn}). Finally, $\gamma$ controls the smoothing strength at a given time. 

\subsection{Link with label smoothing}
A comparison must be drawn with label smoothing~\citep{DBLP:conf/cvpr/SzegedyVISW16} which replaces binary cross-entropy labels $y$ with a smooth version $q$ between 0 and 1. By shifting the optimum from $y$ to $q$, label smoothing prevents models overconfidence, which could improve robustness against the noisy nature of event prediction \citep{DBLP:conf/icml/LukasikBMK20, DBLP:conf/nips/MullerKH19}. Still, despite recent extensions \cite{DBLP:conf/aistats/LiDB20,DBLP:conf/acl/MeisterSC20,DBLP:conf/aaai/LienenH21}, label smoothing remains designed for i.i.d. classification problems. Based on prior knowledge of the temporal structure in our task, our approach also modulates smoothing as a function of time. To the best of our knowledge, we are the first work to introduce a temporal dependence to label smoothing. 

\subsection{Link with multi-horizon prediction}\label{sec:link_mhp}

Temporal label smoothing effectively adapts the contribution of each sample to reflect prior knowledge about the structure of event prediction labels. Under simplifying assumptions justified empirically in Section \ref{sec:Ablations}, we show that MHP can be seen as a special case of temporal label smoothing.  Unlike this method, TLS does not require any architectural change.

\begin{theorem}
\label{prop:MHP}
Under the assumption that model outputs are \textbf{equal} for all horizons $\{h_1,\ldots h_H\}$ (rather than monotonically increasing), MHP is equivalent to temporal label smoothing parameterized with $q^{step}$:\begin{equation*}
\resizebox{0.45\textwidth}{!}{$
    q^{step}(t) = 
    \begin{cases}
    0 & \mathrm{if} \quad t < t_e - h_H \\
    1-\frac{k}{H} & \mathrm{if} \quad  t_e -  h_{k+1} \leq   t < t_e -h_k  \quad \forall k \leq H -1 \\
    1 & \mathrm{if}  \quad  t \geq t_e -h_1
    \end{cases}$}
\end{equation*}
\end{theorem}
\begin{comment}\begin{figure}
    \centering
    \includegraphics[width=0.36 \textwidth]{figures/step_smoothing_time.pdf}
    \caption{\textbf{Label smoothing strength over time}, with staircase parametrization and $(h_{min},h_{max})=(0,2h)$.}
    \label{fig:smoothing_MHP}
\end{figure}
\end{comment}
\begin{proof} See Appendix \ref{appendix:MHP_link}.\end{proof}

Proposition~\ref{prop:MHP} frames MHP as a special case of TLS with parametrization $q^{step}$. This function is defined as a sequence of step functions in time and is illustrated in Figure \ref{fig:smoothing_MHP}.

\section{Experimental setup}\label{sec:exp_setup}
\subsection{Early prediction tasks} \label{sec:tasks}

We demonstrate the effectiveness of our method on different clinical early prediction tasks to understand its added value. These tasks are established in existing literature and published benchmarks and deal with electronic health records from the ICU, where early prediction of organ failure or acute deterioration is critical to patient management~\citep{sutton2020}. Clinical events are labeled following internationally accepted criteria \citep{harutyunyan2019multitask, yeche2021}. 

Our work is first evaluated on the prediction of acute circulatory failure within the next $h=12$ hours, as defined in the HiRID-ICU-Benchmark (HiB)~\citep{yeche2021}. This task is based on the publicly available HiRID dataset~\citep{hyland2020}, containing high-resolution observations of over 33,000 ICU admissions. We also investigate early prediction of patient mortality, or \textit{decompensation}, within a horizon of $h=24$ hours -- a widely studied task in the machine learning literature~\citep{bellamy2020evaluating}. We use the framework defined in the MIMIC-III Benchmark (M3B) \citep{harutyunyan2019multitask} for the MIMIC-III dataset~\citep{johnson2016}, counting approximately 40,000 patient stays. Positive label prevalence is 4.3\% and 2.1\% of time points for circulatory failure and decompensation prediction respectively. Further details on task definition and data pre-processing are provided in Appendix~\ref{appendix:dataset_details}.

\paragraph{Alternative tasks.} To investigate a third clinical event prediction task, we also considered predicting respiratory failure in intensive care patients~\citep{yeche2021}. Unfortunately, ambiguous labeling led to close to random performance for all considered methods. Instead, we benchmarked TLS and baselines on a subtask with better defined labels, prediction of the onset of mechanical ventilation and reached similar conclusions to other tasks in Section~\ref{sec:results}. Experimental details are included in Appendix~\ref{appendix:resp}.

\subsection{Benchmarking strategy}
\paragraph{Baselines.} We quantify the added value of our method by comparing its performance to alternative learning approaches used for early event prediction (EEP) and dynamic survival analysis (DSA), discussed in Section~\ref{sec:pf_relatedwork}. 
Our first baselines consist of {balanced cross-entropy} \citep{king2001logistic} and {focal loss} \citep{lin2017}, popular sample reweighting methods for imbalanced tasks.
We also implement {multi-horizon prediction} as a multi-output model trained to predict event occurrence over different horizons between $0$ and $2h$. Note that for a fair comparison, we set $(h_{min},h_{max})=(0,2h)$ in TLS. As in \citet{tomavsev2019}, a cumulative distribution function layer on logits enforces the monotonicity of predictions (Eq.~\ref{eq:horizon_monotonous}). We also compare to DSA objectives, with landmarking~\citep{van2007dynamic} and the recently proposed TCSR~\citep{Maystre2022}.
Finally, we also compare our method to conventional label smoothing~\citep{DBLP:conf/cvpr/SzegedyVISW16} to confirm that our method's performance can be attributed to its temporal dependency.

\paragraph{Architecture choice.} As our method and baselines are model-agnostic and only vary in terms of optimization objective, a unique model architecture is used for each task, selected through a random search on cross-entropy validation performance. Following a published benchmark on the HiRID dataset \citep{yeche2021}, we use a GRU \citep{DBLP:journals/corr/ChungGCB14} architecture for the {circulatory failure} task. For {decompensation} prediction, transformers \citep{DBLP:conf/nips/VaswaniSPUJGKP17} outperform the LSTM-based models \citep{hochreiter1997long} originally proposed in the M3B benchmark \citep{harutyunyan2019multitask}, and are thus used in our work. As recommended by \citet{tomavsev2019}, we apply $l_1$-regularization to input embedding layers, which improves performance on both tasks. 

Hyperparameters introduced by baselines or by our method, such as strength term~$\gamma$ in smoothing parametrization $q^{exp}$, are optimized through grid searches on the validation set. Further implementation details are provided in Appendix \ref{appendix:implementation_details}.

\subsection{Evaluation metrics}
To account for the highly imbalanced nature of clinical early prediction tasks, the area under the precision-recall curve (AUPRC) provides more insight than the area under the receiver operating characteristic curve (AUROC): under a low prevalence of positive samples, precision is more sensitive to false alarms than specificity \citep{saito2015precision}. Still, "area under the curve" metrics can be poorly representative of clinical usefulness, as improvements in low precision regions can dominate such global metrics but remain incompatible with the low false alarm rates required for clinical deployment. Thus, to better assess model performance in this context, we also measure performance at a clinically motivated operating point through recall at 50\% precision ~\cite{tomavsev2021}. To ensure that conclusions made for this operating point also hold at higher precision constraints, we also plot full precision-recall curves. 

In addition to \textit{timestep-level} metrics, which measure prediction performance at each data point, we also evaluate models in an event-based approach~\cite{hyland2020,tomavsev2019}. Following \citet{tomavsev2019}'s definition, an event prediction is positive if the model outputs a positive prediction at any time over the $h$ hours before the event. The threshold defining a positive prediction is chosen based on a precision lower bound. We also use a stepwise criterion with a 50\% precision. This allows us to measure the event recall of our approach in comparison to published baselines. Unless stated otherwise, we always report mean performance with 95\% confidence intervals on the mean computed over ten training runs.

\begin{figure}[hbt]
    \centering
    \includegraphics[width=0.45\textwidth]{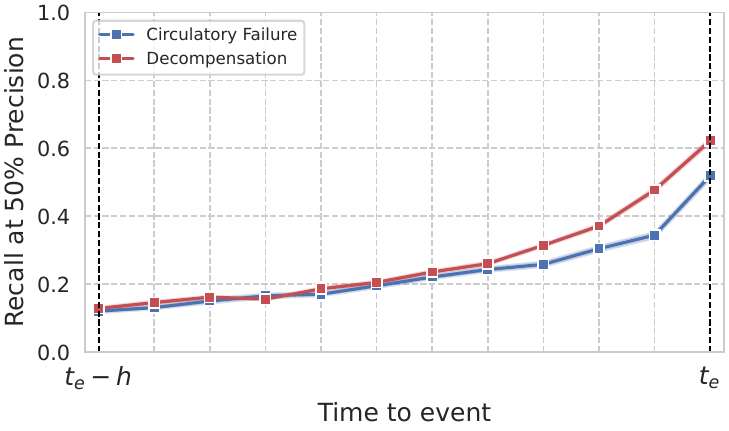}
    \caption{\textbf{Comparison of naive performance as a function of time} on both tasks using cross-entropy. Events should be predicted at a horizon $h$ from an event at time $t_e$. Performance is reported at time increments $\frac{h}{12}$.}
    \label{fig:tasks_perf_v_time}
\end{figure}

\begin{table*}[t] \centering
    \caption{\textbf{Performance of different training objectives for early prediction.} Recall is reported at a 50\% timestep-level precision. {In \textbf{bold}, we highlight best-performing methods with statistically significant $p$-values ($<0.05$) under paired Student's t-tests~\citep{student1908probable}} compared with the next-best method marked italic (last row). Note that cross-entropy is a special case of weighted cross-entropy and focal loss, which performs best in this setting. Hence, the first three lines are identical.}
    \label{tab:perf_results}
\resizebox{\textwidth}{!}{\begin{tabular}{lcccccc}
\toprule

 Task & \multicolumn{3}{c}{Circulatory Failure (HiRID)} & \multicolumn{3}{c}{Decompensation (MIMIC-III)} \\
 \cmidrule(lr){2-4} \cmidrule(lr){5-7}
Training objective &         AUPRC & Timestep  Recall & Event Recall &        AUPRC &  Timestep Recall &Event Recall \\
\midrule
Cross-entropy   \citep{Lauritsen2020, hyland2020}     &             39.1 $\pm$ 0.4 &             29.3 $\pm$ 0.9 &   82.8 $\pm$ 1.3 &        34.5 $\pm$ 0.4 &             28.2 $\pm$ 0.5 & 69.7 $\pm$ 1.0\\
Weighted CE  \cite{king2001logistic} &             39.1 $\pm$ 0.4 &             29.3 $\pm$ 0.9 &  82.8 $\pm$ 1.3  &         34.5 $\pm$ 0.4 &             28.2 $\pm$ 0.5 & 69.7 $\pm$ 1.0 \\
Focal loss  \cite{lin2017,wang2020feature, roy2022disability} &             39.1 $\pm$ 0.4 &             29.3 $\pm$ 0.9 & 82.8 $\pm$ 1.3 &            34.5 $\pm$ 0.4 &             28.2 $\pm$ 0.5 & 69.7 $\pm$ 1.0\\
Label smoothing \citep{DBLP:conf/cvpr/SzegedyVISW16}      &             39.3 $\pm$ 0.4 &             29.9 $\pm$ 0.8 &     83.8 $\pm$ 1.3 &       33.9 $\pm$ 0.3 &             27.7 $\pm$ 0.5 & 68.8 $\pm$ 1.0 \\
Multi-horizon \citep{tomavsev2019,jarrett2019dynamic} &             {\it 39.6} $\pm$ 0.5 &             {\it 30.3} $\pm$ 1.0 & 85.2 $\pm$ 1.7      &      {\it 34.9} $\pm$ 0.3 &             {\it 28.6} $\pm$ 0.5 & {\it 70.3} $\pm$ 0.6 \\\midrule
 Landmarking \cite{van2007dynamic,parast2014landmark} &             {\it 39.6} $\pm$ 0.3 &           30.1 $\pm$ 0.6 & {\it 89.1} $\pm$ 0.8&   34.0 $\pm$ 0.5 &           27.2 $\pm$ 0.6 &  68.8 $\pm$ 1.1 \\
 TCSR \cite{Maystre2022} &         36.0  $\pm$ 0.4  &	26.5  $\pm$ 0.8  &  89.0 $\pm$ 2.1&  28.6 $\pm$ 1.2 &           19.9 $\pm$ 1.4 & 68.4 $\pm$ 1.0\\
\midrule
\textbf{Temporal label smoothing}     &  $\mathbf{40.6}$ $\pm$ 0.3 &  $\mathbf{32.3}$ $\pm$ 0.7 & \textbf{92.5} $\pm$ 0.5 &$\mathbf{35.5}$ $\pm$ 0.3 &  $\mathbf{29.3}$ $\pm$ 0.4 & \textbf{71.8} $\pm$ 0.8\\
\midrule
{$p$-value}  & \textbf{0.002} & \textbf{0.004}  &  \textbf{<0.001} & \textbf{0.004} & {\textbf{0.02}}  &  \textbf{0.002}\\
\bottomrule
\end{tabular}}
\end{table*}

\section{Results}\label{sec:results}

In this section, we validate the following claims: (1) temporal label smoothing yields practical performance improvement along clinically-motivated metrics, and (2) achieves this by leveraging temporal structure and modulating prediction confidence as a function of event proximity.

\subsection{Prediction performance}\label{sec:perf-results}

Overall, our results highlight that TLS improves performance over other approaches proposed to address the challenges of early clinical prediction. We occasionally focus on circulatory failure prediction for brevity; see Appendix~\ref{appendix:add_exp} for similar conclusions on decompensation.

\paragraph{Necessity of temporal inductive biases.} As visualized in Figure~\ref{fig:tasks_perf_v_time}, training EEP as a simple binary classification with a cross-entropy objective shows a reduction in recall between event time $t_e$ and prediction horizon $t_e - h$. This suggests a weakening in the discriminative signal associated with events and an increase in noise close to the label boundary, where performance is the poorest. In fact, we argue that correct predictions in this region, close to $t_e-h$, are not as critical as ones near $t_e$: missing an imminent event is more severe. Mirroring the decrease in both signal strength and clinical importance of predictions as the time-to-event increases, model confidence should also decrease, focusing instead on more critical time windows.

\begin{figure}[h]
\centering
    \centering
    \includegraphics[width=0.9\linewidth]{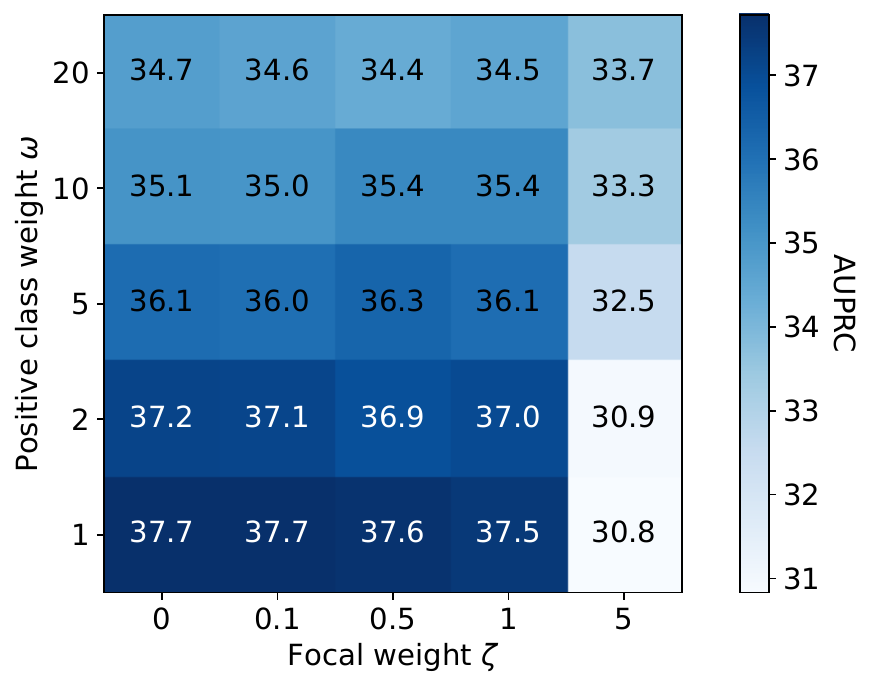}
\caption{\textbf{Performance loss with class reweighting methods} on circulatory failure prediction (validation). Balanced cross-entropy corresponds to $\zeta=0$.}
\label{fig:focal+weighted}
\end{figure}

\begin{figure*}[t]
\centering
\begin{subfigure}[b]{0.46\textwidth}
  \centering
  \includegraphics[width=\linewidth]{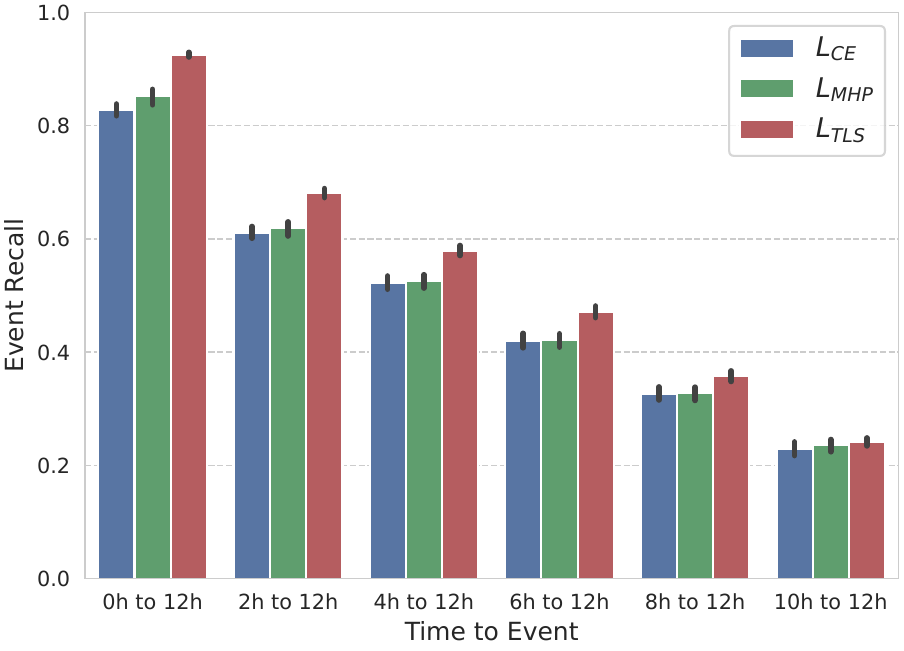}
  \caption{\centering Event-level performance for a 50\% timestep-level precision threshold.} \label{fig:event-based}
\end{subfigure}
\begin{subfigure}[b]{0.46\textwidth}
 \centering
  \includegraphics[width=\linewidth]{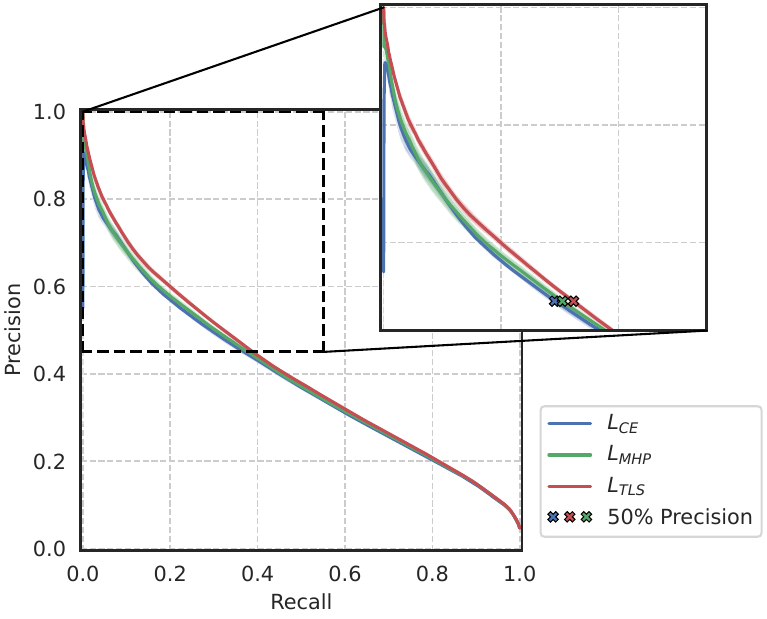}
  \caption{\centering Precision-recall curve. Inset shows the clinically-applicable region with precision $>50$\%.}
  \label{fig:PR_curve}
\end{subfigure}
\caption{\textbf{Clinically-oriented performance analysis} of different training objectives on circulatory failure prediction (CE: cross-entropy, MHP: multi-horizon prediction, TLS: temporal label smoothing).}

\label{fig:clinical_performance_circ}
\end{figure*}

\paragraph{Timestep-level performance.} In Table \ref{tab:perf_results}, we find TLS to outperform baselines across all metrics for circulatory failure and decompensation\footnote{Despite overlapping confidence intervals between multi-horizon and TLS on decompensation due to individual training run variability, we can reject the null hypothesis that MHP has a higher performance than our method ($p$-values $< 0.05$)}. The full precision-recall curve of models trained with the best objectives is shown in Figure \ref{fig:PR_curve}: TLS improves recall for all precision thresholds beyond 50\%, a low false-alarm region of particular clinical relevance~\citep{sutton2020}.

In contrast, loss reweighting methods designed to tackle class imbalance were found to reduce performance on all tasks over traditional cross-entropy, as shown in Figure~\ref{fig:focal+weighted}. For weighted cross-entropy, we attribute it to the increase in false alarms resulting from the drive to improve recall. This further reduces the low precision of all models, thus negatively affecting the AUPRC. On the other hand, focal loss down-weighs confident samples in training, constraining the model to focus on samples with uncertain predictions. In the context of noisy labeling, as is the case close to our class boundary, data points with ambiguous signals cannot be correctly predicted and thus dominate the loss, impeding improvements in other regions of input space. We analyze model performance over time in Section~\ref{sec:Ablations} to further support this hypothesis.

\paragraph{Empirical comparison to dynamic survival analysis.} Despite the similarities between the tasks of early event prediction and dynamic survival analysis, survival objectives were not found to markedly improve performance on the former, as shown in the second block of Table \ref{tab:perf_results}. A likely explanation for this is that the survival likelihood is trained to predict events potentially occurring at horizons much greater than that of interest in EEP. As signal strength decreases with the time-to-event, errors from distant events dominate the loss -- leading to poor performance on long time series. This finding goes in the direction of recent works \cite{jarrett2019dynamic,lee2019dynamic} in dynamic survival analysis, which train a fixed (multi-)horizon model as in EEP.

Finally, the prediction-dependent label smoothing in TCSR~\citep{Maystre2022}, designed to improve survival performance on short-sequence survival tasks, did not improve performance our EEP tasks either. Training was found to be unstable due to error propagation over long sequences.

\paragraph{Clinically relevant performance.} As highlighted in Figure \ref{fig:event-based}, TLS improves performance over other training objectives in predicting overall adverse event episodes throughout a stay. For circulatory failure, temporal label smoothing is able to predict 7.4\% more events than the closest baseline designed for EEP (multi-horizon prediction): this corresponds to reducing the number of missed events by a factor of 2, from 303 to 152 out of 2045 events in the test set on average. Within the events captured by TLS but not by MHP, models trained with our objective predict them on average 104 minutes before their occurrence, giving clinicians sufficient time to take action and avoid patient degradation. We also note here the benefit of adapting dynamic survival analysis to the EEP setting, with landmarking and TCSR performing best in circulatory failure event recall, after TLS. As these methods also enforce temporal structure, this result further motivates our approach, which achieves even greater performance gains, and suggests promise in using survival likelihood objectives for early event prediction.

\subsection{Illustrative insights}
\label{sec:Ablations}

We propose ablations to build intuition around our proposed method. In particular, we aim to understand how temporal smoothing works and why it outperforms other training approaches for early prediction tasks.

\begin{table*}[t]
    \centering
    \caption{\textbf{Do MHP's multiple outputs improve performance over TLS with $q^{step}$?} We provide $p$-values for the paired Student-t test \citep{student1908probable} on the null hypothesis $H_0$:MHP $\leq$ TLS. With no statistically significant improvements ($p<0.05$), we justify our assumption in Proposition~\ref{prop:MHP}.} \label{tab:ablation_MHP_step}
\footnotesize
\begin{tabular}{lcccccc}
\toprule
Task & \multicolumn{3}{c}{Circulatory Failure (HiRID)} & \multicolumn{3}{c}{Decompensation (MIMIC-III)} \\
 \cmidrule(lr){2-4} \cmidrule(lr){5-7}
Training objective &         AUPRC & Timestep  Recall & Event Recall &        AUPRC &  Timestep Recall &Event Recall \\
\midrule
MHP &             39.6 $\pm$ 0.5 &             30.3 $\pm$ 1.0 & 85.2 $\pm$ 1.7      &      34.9 $\pm$ 0.3 &            28.6 $\pm$ 0.5 & 70.3 $\pm$ 0.6  \\
TLS ($q^{step}$)                  &             39.3 $\pm$ 0.2 &             29.4 $\pm$ 0.8 & 
83.4 $\pm$ 1.2 &         35.2 $\pm$ 0.3 &             29.2 $\pm$ 0.4 &      70.4 $\pm$ 0.7    \\ \midrule
p-value ($H_0$)     &                       0.11 &                       0.10 &      \textbf{0.03}  &             0.95 &                       0.97 &      0.40 \\
\bottomrule
\end{tabular}\end{table*}

\paragraph{Empirical comparison to multi-horizon prediction.} In our theoretical discussion in Section \ref{sec:link_mhp}, we demonstrated how MHP is a restriction of label smoothing with a step function $q^{step}(t)$. This claim relies on the constraint to produce a unique prediction across all considered horizons, reflecting the design of our method. We verify the impact of this assumption by measuring performance gains afforded by learning distinct predictions per horizon. As shown in Table~\ref{tab:ablation_MHP_step}, we only find statistical evidence for slight performance gain over using $q^{step}$ on event recall for circulatory failure. Thus, models do not appear to leverage this additional flexibility offered by MHP. With superior results on all event- and timestep-based experiments, and a simpler implementation, we find temporal label smoothing to be a superior training objective to MHP in early prediction tasks.

\begin{figure}[h]
\begin{subfigure}[b]{0.47\textwidth}
  \centering
  \includegraphics[width=\linewidth]{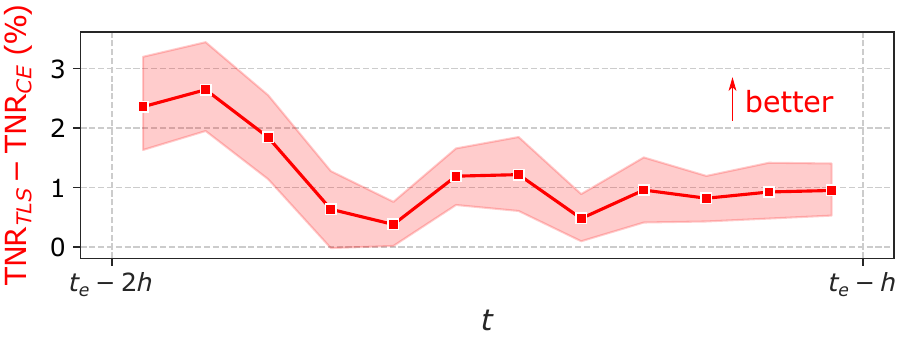}\caption{True negative rate (TNR).}
  \label{fig:delta_tnr_circ}
\end{subfigure} \hfill
\begin{subfigure}[b]{0.47\textwidth}
 \centering
  \includegraphics[width=\linewidth]{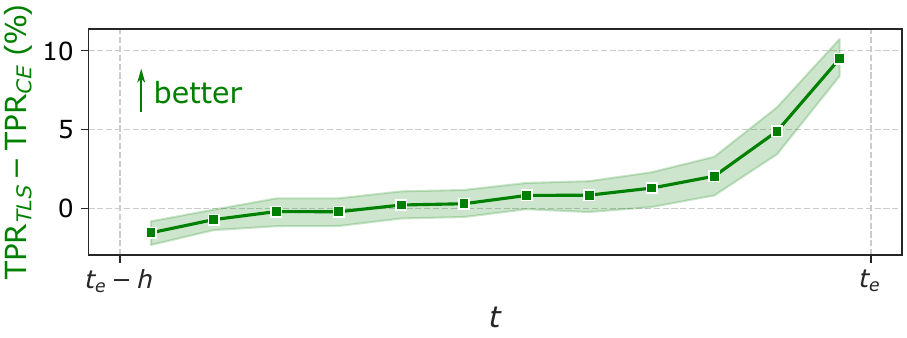}\caption{True positive rate (TPR).}
  \label{fig:delta_tpr_circ}
\end{subfigure}
\caption{\textbf{Performance improvement over time} for TLS over cross-entropy on circulatory failure.
Timestep-level metrics computed for precision of $0.5$ over two-hour bins.}
\label{fig:plot_delta}
\end{figure}

\paragraph{Performance over time.} To better understand the mechanism of action of TLS, we study the difference in performance between TLS and the cross-entropy objective over time in Figure~\ref{fig:plot_delta}. TLS results in a significant increase in true positive and negative rates when prediction time is far from the label boundary ($t=t_e-2h$ or $t=t_e$). In particular, the performance gains close to the event time $t_e$ explains the better recall of imminent events in Figure~\ref{fig:event-based}.

In contrast, the prediction model trained with TLS is less competitive where smoothing is strongest, near $t_e-h$, but, as expected, this performance loss remains minor. This result validates our hypothesis that the signal is too noisy in the boundary region for any model to recover the original label distribution. From a clinical perspective, errors made in the boundary region are less critical, as they result in the latest false positives or earliest false negatives. Overall, TLS not only improves global event prediction performance but allows these gains to occur at more critical times for clinicians.

\section{Conclusion}

Early event prediction is paramount to the development of clinical decision support systems, with a demonstrated potential to improve patient outcomes \citep{smith2013}. Still, this task remains relatively poorly studied in the machine learning literature, with few training solutions tailored to address its challenges or to exploit its intrinsic temporal structure. We demonstrate that this can be achieved by adapting and significantly improving approaches from the survival analysis literature~\citep{van2007dynamic, parast2014landmark}. This also motivates us to design a simple, yet top-performing training framework that leverages the structure of event signals over time. We show that multi-horizon prediction, a heuristic used to improve early prediction, can be formalized as a realization of our framework. 

{\it Temporal label smoothing empirically outperforms all considered baselines} on various tasks and datasets, {\it with significant improvements} in clinically-relevant evaluation metrics. Our ablation studies show that it effectively focuses training on data points with a stronger predictive signal.

Promising avenues of further work include combining TLS with survival objectives and using temporal label smoothing for survival regression tasks, to explore the relative benefits of different temporal inductive biases. Looking ahead, we expect that temporal label smoothing will be leveraged to develop more clinically reliable systems for risk prediction of infrequent adverse events. Further research on tailored machine learning solutions to improve real-world decision support holds promise for better clinical care and operations management.

\newpage
\bibliographystyle{unsrtnat}
\bibliography{references}

\appendix
\newpage
\onecolumn

\section{Theoretical details} 

\subsection{Multi-Horizon prediction: proof of Proposition \ref{prop:MHP}}
\label{appendix:MHP_link}
\paragraph{Equivalency between MHP and TLS objectives.}
Recalling the formalism of multi-horizon prediction outlined in Section~\ref{sec:link_mhp}, true labels and model predictions at time $t$ can be rewritten as $\mathbf{y}_{t} = [y^{h_1}_{t}, \ldots, y^{h}_{t}, \ldots, y^{h_H}_{t}]$ and $\hat{\mathbf{y}}_{t} = [\hat{y}^{h_1}_{t}, \ldots, \hat{y}^{h}_{t}, \ldots, \hat{y}^{h_H}_{t}]$, where $H$ is the number of horizons considered. The training objective for this datapoint becomes:
    \begin{equation*}
            L^{MHP}(\mathbf{y}_{t},\hat{\mathbf{y}}_{t}) = -\frac{1}{H}\sum_{k=1}^H y^{h_k}_{t}\log(\hat{y}^{h_k}_{t}) + (1-y^{h_k}_{t})\log(1 - \hat{y}^{h_k}_{t})
    \end{equation*}

The assumption that  $\{\hat{y}^{h_k}_{t}\}_{k}$ is \textbf{equal} for all $k$ allows to rewrite the objective as follows:
    \begin{equation*}
             L^{MHP}(\mathbf{y}_{t},\hat{\mathbf{y}}_{t}) = -\left[\log(\hat{y}_{t})\frac{1}{H}\sum_{k=1}^H y^{h_k}_{t} + \log(1 - \hat{y}_{t})\frac{1}{H}\sum_{k=1}^H (1-y^{h_k}_{t})\right]
    \end{equation*}
with $\hat{y}_{t}$ being the common prediction shared across all horizons. This equation can now be viewed as a temporal label smoothing objective with smoothed labels $q^{step}(t) = \frac{1}{H}\sum_{k=1}^H y^{h_k}_{t}$:
    \begin{equation*}
             L^{MHP}(\mathbf{y}_{t},\hat{\mathbf{y}}_{t}) = -\left[\log(\hat{y}_{t}) \cdot q^{step}(t) + \log(1 - \hat{y}_{t})\cdot \left( 1-q^{step}(t) \right) \right]
    \end{equation*}
    
\paragraph{Smoothing parametrization.}
Next, we aim to recover the explicit form of $q^{step}(t)$. Without loss of generality, we assume that horizons $\{h_k\}_k$ are in ascending order. The temporal dependency between samples, formalized in Equation~\ref{eq:time_monotonous}), results in the following relationship between predictions at horizons $h_u$ and $h_v$ :
\begin{align} \label{eq: mhp_predictions_order_1}
    & v \leq u \quad \text{and} \quad y^{h_v}_{t} = 1 \implies  y^{h_u}_{t} = 1\\
    & v \geq u  \quad \text{and} \quad y^{h_v}_{t} = 0 \implies  y^{h_u}_{t} = 0  \label{eq: mhp_predictions_order_0}
\end{align}

\begin{figure}[h]
    \centering
    \includegraphics[width =0.9\linewidth]{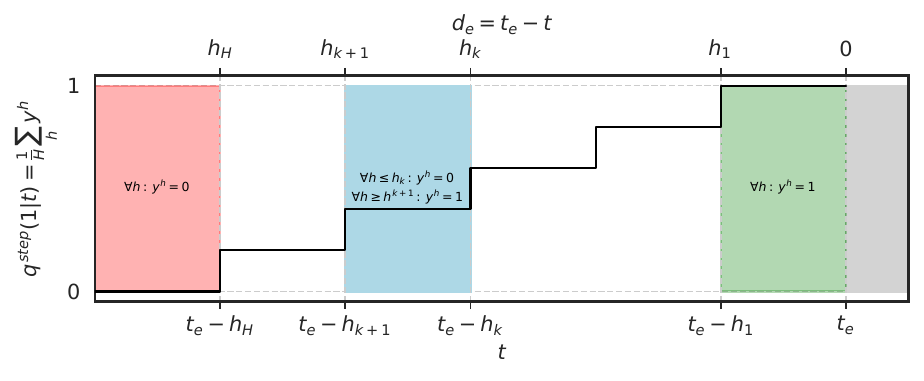}
    \caption{\textbf{Label values for multi-horizon prediction}, and conversion to smoothed labels $q^{step}(t)$.}
    \label{fig:appendix_MHP}
\end{figure}

Thanks to the above property, we can determine $q^{step}(t)$ by studying three cases of multi-horizon labels, illustrated in Figure \ref{fig:appendix_MHP}. For notational simplicity, we define the time-to-event as $d_e( t) = t_e - t$.

\definecolor{lightgreen}{HTML}{2ca02c}
\definecolor{tabblue}{HTML}{1f77b4}
\definecolor{tabred}{HTML}{d62728}
\textcolor{lightgreen}{\textbf{Case 1: $d_e( t) \leq h_1$}.}\\
From label definition. we have that $ y^{h_1}_{t} = 1$ if $d_e( t) \leq h_1$. As $h_1$ is the smallest horizon, following Equation~\ref{eq: mhp_predictions_order_1}, we have $ y^{h_c}_{t} = 1, \forall c \in \llbracket 1, H \rrbracket $. We can rewrite the objective as:
    \begin{align*}
             L^{MHP}(\mathbf{y}_{t},\hat{\mathbf{y}}_{t}) &= -\log(\hat{y}_{t})\\
                      &= -[q^{step}(t)\log(\hat{y}_{t}) + (1 - q^{step}(t))\log(1 - \hat{y}_{t})]
    \end{align*}
where $q^{step}(t) = 1$.

\textcolor{tabred}{\textbf{Case 2: $d_e( t)  >  h_H$}.} \\
Similarly, if $d_e( t)  >  h_H$, then $y^{h_H}_{t} = 0$ which implies $ y^{h_c}_{t} = 0, \forall c \in \llbracket 1, H \rrbracket $ from Equation~\ref{eq: mhp_predictions_order_0}. The objective can be rewritten as:
    \begin{align*}
             L^{MHP}(\mathbf{y}_{t},\hat{\mathbf{y}}_{t}) &= -\log(1-\hat{y}_{t})\\
                      &= -[q^{step}(t)\log(\hat{y}_{t}) + (1 - q^{step}(t))\log(1 - \hat{y}_{t})]
    \end{align*}
where $q^{step}(t) = 0$.

\textcolor{tabblue}{\textbf{Case 3: $\exists k \in \llbracket 1, H-1 \rrbracket \quad \mathrm{s.t} \quad h_k < d_e(t)  \leq   h_{k+1}$}.}\\
Following the same reasoning as in the first two cases, we now have a specific index $k$ which separates positive and negative labels. We have $y^{h_c}_{t} = 0, \forall c \in \llbracket 1, k \rrbracket$ and $y^{h_c}_{t} = 1, \forall c \in \llbracket k+1, H \rrbracket$. This allows to rewrite the objective as follows:
    \begin{align*}
            L^{MHP}(\mathbf{y}_{t},\hat{\mathbf{y}}_{t}) &= -[\frac{H-k}{H}\log(\hat{y}_{t}) + \frac{k}{H}\log(1-\hat{y}_{t})]\\
                      &= -[q^{step}(t)\log(\hat{y}_{t}) + (1 - q^{step}(t))\log(1 - \hat{y}_{t})]
    \end{align*}
where
\begin{equation*}
    q^{step}(t) = \frac{H-k}{H}.
\end{equation*}

This defines a new smoothing parametrisation $q^{step}$: \begin{equation*}
    q^{step}( t) = 
    \begin{cases}
    1-\frac{k}{H} & \mathrm{if} \quad  h_k \leq  d_e(t) < h_{k+1} \quad \forall k \leq H -1 \\
    1 & \mathrm{if}  \quad d_e(t) \leq h_1 \\     
    0 & \mathrm{if} \quad d_e(t) > h_H \\
    \end{cases}
\end{equation*}

Thus, $\forall d_e(t) > 0$, we find that $L^{MHP} = L^{TLS}$ when smoothed labels are defined as $q^{step}$. This concludes our proof. \hfill \qedsymbol

\subsection{Temporal label smoothing functions}
\label{appendix:temporal_smoothing_fn}

\begin{figure}[h]
\centering
\begin{subfigure}{0.49\textwidth}
  \centering
  \includegraphics[width=\linewidth]{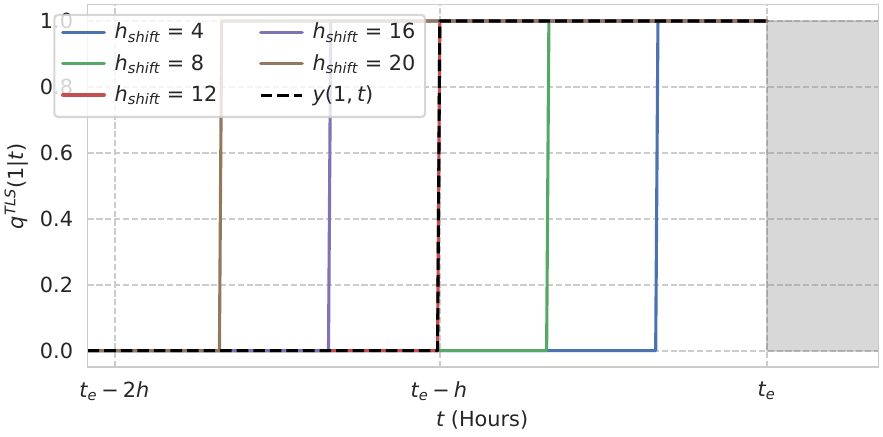}
  \caption{\centering {$q^{shift}$}} \label{fig:qshift}
\end{subfigure}
\hspace{2pt}
\begin{subfigure}{0.49\textwidth}
  \centering
  \includegraphics[width=\linewidth]{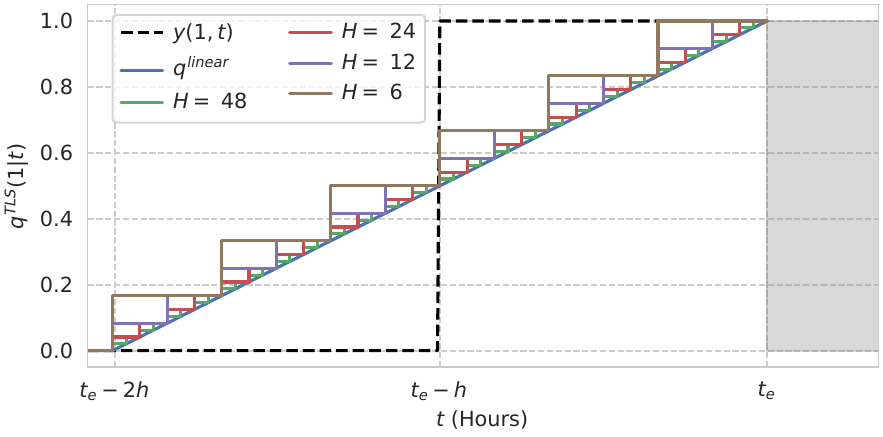}
  \caption{\centering $q^{step}$ and $q^{linear}$}
  \label{fig:step_TLS}
\end{subfigure}
\vspace{0.5em}

\begin{subfigure}{0.49\textwidth}
  \centering
  \includegraphics[width=\linewidth]{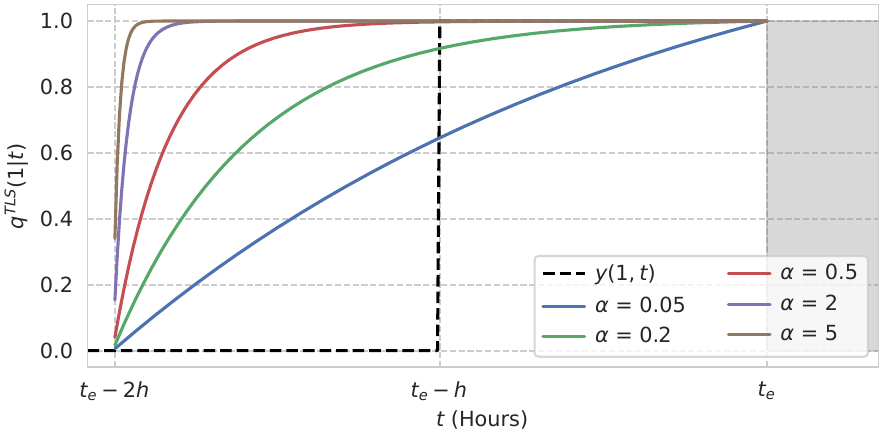}
  \caption{{\centering $q^{concave}$}} \label{fig:concave}
\end{subfigure}
\hspace{2pt}
\begin{subfigure}{0.49\textwidth}
  \centering
  \includegraphics[width=\linewidth]{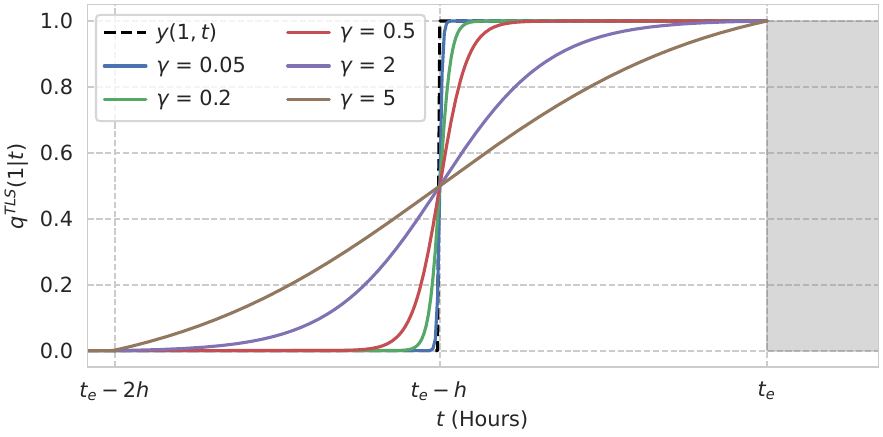}
  \caption{\centering \centering $q^{sigmoid}$}
  \label{fig:sigmoid_TLS}
\end{subfigure}
\caption{\textbf{Illustration of temporal label smoothing} with alternative smoothing parametrizations.} \label{fig:TLS_other}
\end{figure}

\begin{figure}[h]
\centering
\begin{subfigure}[b]{0.9\textwidth}
  \centering
  \includegraphics[width=\linewidth]{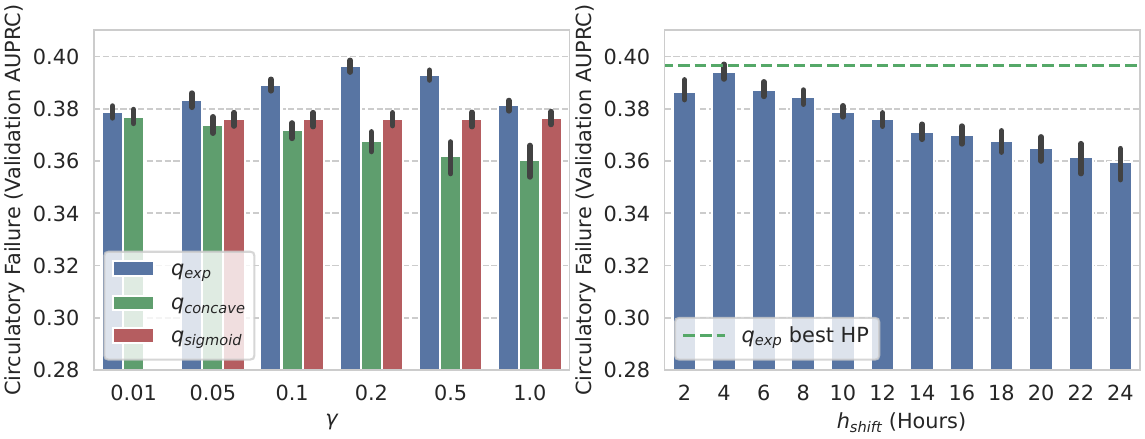}
  \caption{\centering {Circulatory failure.}} 
  \label{fig:smoothing_hypers_circ}
\end{subfigure}
\vspace{0.5em}
\begin{subfigure}[b]{0.9\textwidth}
  \centering
    \includegraphics[width=\linewidth]{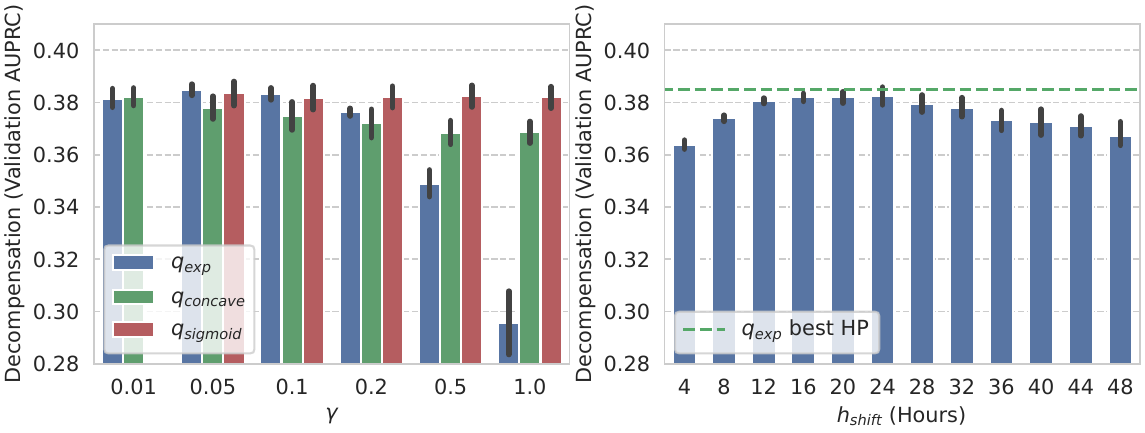}
  \caption{\centering {Decompensation.}} 
  \label{fig:smoothing_hypers_decomp}
\end{subfigure}
\caption{{\textbf{Validation AUPRC performance of temporal label smoothing as a function of smoothing hyperparameters}, with different smoothing parameterizations. (Left) Performance for different smoothing strengths $\gamma$ with $q^{exp},q^{concave},q^{sigmoid}$; (Right)  Performance for different prediction horizons $h_{shift}$ with $q^{shift}$ smoothing.}} \label{fig:smoothing_hyperparams}
\end{figure}

Motivated by prior work \citep{tomavsev2019, cox1972}, we compare the performance of various smoothing functions $ q(t)$. All proposed parametrizations are continuous and monotonous increasing functions that satisfy boundary conditions $q( t_e - 2h) = 0$ and $q( t_e)=1$. As evidenced in Table \ref{tab:other_fn}, we find exponential label smoothing to perform best or as well as others across all tasks and metrics. {Performance as a function of hyperparameter setting can be visualized in Figure \ref{fig:smoothing_hyperparams}. All model and hyperparameter selection were carried out on the validation set, including the final choice of parametrization function.
}

\begin{comment}
    \vspace{0.5em}
\begin{subfigure}[b]{0.9\textwidth}
  \centering
    \includegraphics[width=\linewidth]{figures/hp_search_resp.pdf}
  \caption{\centering {Respiratory failure.}} 
  \label{fig:smoothing_hypers_resp}
\end{subfigure}
\end{comment}

\begin{table}[h] \centering
    \caption{\textbf{Performance of different smoothing functions on early prediction tasks.} Timestep-level recall is reported at a 50\% precision.} \label{tab:other_fn}
\begin{tabular}{lcccc}
\toprule

 Task & \multicolumn{2}{c}{Circulatory Failure} & \multicolumn{2}{c}{Decompensation} \\
 \cmidrule(lr){2-3} \cmidrule(lr){4-5}
Method &         AUPRC &   Recall &         AUPRC &   Recall \\
\midrule
$q^{step}$              &             39.3 $\pm$ 0.2 &             29.4 $\pm$ 0.8 &             35.2 $\pm$ 0.3 &             29.2 $\pm$ 0.4  \\
{$q^{shift}$} & {$40.1 \pm 0.3$} & {$31.8 \pm 0.6$} & {34.5 $\pm$ 0.4} &  {28.2 $\pm$ 0.5}  \\
$q^{linear}$      &  39.4 $\pm$ 0.3 &             29.7 $\pm$ 0.8 &             35.1 $\pm$ 0.4 &             29.2 $\pm$ 0.6  \\
$q^{sigmoid}$     & 39.4 $\pm$ 0.3 &             29.7 $\pm$ 0.8 &             34.9 $\pm$ 0.4 &             28.8 $\pm$ 0.5 \\
{$q^{concave}$}      &  {39.4 $\pm$ 0.3} &             {29.7 $\pm$ 0.8} &             {35.1 $\pm$ 0.4} &             {29.2 $\pm$ 0.6}  \\
$q^{exp}$   &  $\mathbf{40.6}$ $\pm$ 0.3 &  $\mathbf{32.3}$ $\pm$ 0.7 &  $\mathbf{35.5}$ $\pm$ 0.3 &  $\mathbf{29.3}$ $\pm$ 0.4  \\
\bottomrule
\end{tabular}\end{table}

{
\paragraph{Shifted boundary labels.}
Shifting the prediction horizon or label boundary in training can be viewed as a form of temporal label smoothing, in which class labels are inverted within a prediction window of interest. This defines the following smoothing parametrization $q^{shift}(t)$:
\begin{equation}
    q^{shift}( t) = \mathds{1}\left[ t \geq t_e - h_{shift} \right]
\end{equation}
where $h_{shift}$ is a hyperparameter controlling the horizon of the smoothed labels ($h_{shift}=h$ corresponds to cross-entropy training). The strength of this smoothing function is illustrated in Figure~\ref{fig:qshift}.

Figure \ref{fig:smoothing_hyperparams} outlines the performance of this alternative smoothing parametrization as a function of $h_{shift}$. For decompensation, shifting the label boundary closer to the event time decreases performance. On circulatory failure, performance does improve over traditional cross-entropy training as the label horizon is brought closer to the event of interest, which can be interpreted as an inductive bias similar to that induced by the exponential smoothing function.

}

\paragraph{Linear label smoothing.}
The most straightforward extension to the step function $q^{step}$ described in Section~\ref{sec:link_mhp} is a linear label smoothing corresponding to the case $H \rightarrow +\infty$.\\
Our parametrization $q^{linear}( t)$ is thus defined as follows:
\begin{equation}
    q^{linear}( t) = 
    \begin{cases}
    0 & \mathrm{if} \quad t \leq  t_e - 2h \\
    1 - \frac{t_e - t}{2h}& \mathrm{if} \quad t > t_e - 2h    
    \end{cases}
\end{equation}
We illustrate the impact of the number of steps $H$ in Figure~\ref{fig:step_TLS}.

\paragraph{Sigmoidal label smoothing.}
Another natural direction to explore is to smooth labels starting from the true distribution, a unique step function at $t = t_e - h$. This can be achieved by defining $q_t$ as a generalized logistic function \citep{richards1959flexible}:

\begin{equation}
    q^{sigmoid}( t)  = \begin{cases}
    0 & \mathrm{if} \quad t \leq  t_e - 2h \\
    \frac{K-A}{1 + e^{\frac{t_e - t - d}{\gamma}}} + A & \mathrm{if} \quad t > t_e - 2h
    \end{cases}
\end{equation}

where $K$, $A$ and $d$ are three constants fixed by imposing the boundary conditions at $t=t_e - 2h $ and $t=t_e$, as well as $q(t_e - 2h) = \frac{1}{2}$. This yields:
\begin{align*}
K &= -Ae^\frac{2h - d}{\gamma}\\
A &= \frac{e^\frac{- d}{\gamma} + 1 }{e^\frac{- d}{\gamma} - e^\frac{2h - d}{\gamma}}\\
d &= h
\end{align*}

As shown in Figure~\ref{fig:sigmoid_TLS}, $\gamma$ controls the smoothing strength, interpolating between the true distribution $\delta_{y=1}$\ as $\gamma \rightarrow 0$ and $q^{linear}$ when $\gamma \rightarrow +\infty$.

\paragraph{Exponential label smoothing.}
The smoothing function we find to perform best is the exponential decay one. This idea is motivated by survival analysis, where patient survival probability can be modeled as the exponential decay of a cumulative hazard function \cite{cox1972, collett2015modelling}. In practice, as defined in Section~\ref{sec:TLS}, our exponential smoothing function $ q^{exp}( t)$ is defined as follows:
\begin{equation}
    q^{exp}( t)  = \begin{cases}
    0 & \mathrm{if} \quad t \leq  t_e - 2h \\
    e^{ - \gamma ( t_e - t - d) } + A  & \mathrm{if} \quad t > t_e - 2h 
    \end{cases}
\end{equation}

where parameters $\{d, A\}$ are set to satisfy boundary conditions:
\begin{align*}
    A &=  - e^{- \gamma(2h - d)} \\
    d &=  - \frac{1}{\gamma} \ln \left(1 - e^{-\gamma 2h} \right) 
\end{align*}

Here, $\gamma$ also controls the smoothing strength between $q^{linear}$ when $\gamma \rightarrow 0$ and $q(t)=0 \: \forall t < t_e $ when $\gamma \rightarrow +\infty$.

{Overall, despite $q^{shift}$ achieving good results on circulatory failure, $q^{exp}$ statistically outperforms this smoothing parameterization for both tasks on validation metrics. An interesting avenue for further work would be to combine exponential smoothing with the boundary shift approach, or effectively change $(h_{min}, h_{max})$, which was fixed to $(0,2h)$ in our work for a fair comparison to multi-horizon prediction.}

\paragraph{Concave exponential label smoothing.}

Finally, to mirror the behavior of the exponential smoothing function away from linear interpolation and investigate its effect on performance, we designed the following concave smoothing function $q^{concave}$:
\begin{equation}
    q^{concave}( t)  =  
    \begin{cases}
    0 & \mathrm{if} \quad t \leq  t_e - 2h \\
    1 - e^{ - \gamma (d - t_e + t) } + A & \mathrm{if} \quad t > t_e - 2h
    \end{cases}
\end{equation}
Parameters $\{d, A\}$ are identical to the convex smoothing function parameters, set to satisfy boundary conditions. The strength of this concave smoothing function is illustrated Figure \ref{fig:concave}.

No performance gains were obtained through temporal label smoothing with a concave function, as shown in Figure \ref{fig:smoothing_hyperparams}. This smoothing function effectively penalizes false positives harder than false negatives, which is less adapted to our tasks of interest (in contrast to the convex $q^{exp}$). As a result, the best-performing concave parametrization is consistently obtained with the lowest value of $\gamma$, closer to a linear function choice.

\subsection{Related time-series tasks} \label{appendix:related_tasks}

\begin{comment}

\paragraph{Comparison to survival analysis.} 
\TODO{}
Survival analysis consists of statistical methods concerned with predicting the probability of a certain event taking place over time \citep{collett2015modelling}. In our formalism outlined in Section \ref{sec:problem_formalism}, the corresponding task is to regress the time of the next event $t_e$ based on patient information accumulated up to time $t$. To recover early event prediction, a threshold on the hazard model can thus be applied to determine whether an event will happen within our horizon of interest $h$. Modeling constraints imposed in survival analysis improve time-to-event prediction performance over traditional regression methods, which supports our approach to leverage the temporal structure of our comparable task. Interestingly, recent developments in survival modeling to deal with dynamic predictions have been addressed with multi-horizon prediction \citep{jarrett2019dynamic}.

Still, distinctions must be highlighted between our adverse event prediction problem and the typical experimental setup for survival analysis: in our case, multiple events can occur over the course of a patient's stay, with unknown patient states during and immediately after event occurrence. This results in complex, informative censoring patterns and challenges common assumptions in survival analysis, which can therefore not be directly applied to our task.
\end{comment}

{ \paragraph{Comparison to early time-series classification.} 

A distinction must be drawn between our task of early event prediction and that of early time-series classification. The latter has been more extensively explored in the literature \citep{Xing2009,He2013,Yang2021dir}, but addresses a distinct problem. 

Considering a time series up to timestep $t$, early event prediction is concerned with classifying \textit{whether} a particular event will occur between $t$ and $t+h$, for a fixed horizon $h$. Predictions are made at each timepoint over the entire time series: as multiple samples arise from the same time series and therefore depend on one another over time, these should not be considered as i.i.d.

In contrast, early classification of time series aims to regress \textit{the first timepoint} $t$ at which a label for the entire time series can be predicted with a desired accuracy~\citep{Xing2009}. A single prediction is made, as soon as possible, for the entire series -- which can be considered an independent sample from the dataset of time series. This latter task can be framed as an early prediction of the event “prediction is possible”, where $h=\infty$, given a separate time-series classifier. As a result, an interesting avenue of further work would be to apply temporal label smoothing to the latter task.

On the other hand, early event prediction cannot be translated into a simple early classification problem. As a result, methods designed for early time-series classification are therefore not applicable to this problem setting. 

}

\subsection{EEP Objective Functions} \label{appendix:relatedwork}

In this section, we clarify the mathematical formalism behind our EEP baselines to facilitate comparison to temporal label smoothing. Most baselines explored effectively propose a modification of the cross-entropy objective often used for binary classification tasks, $L^{CE}(y,\hat{y}) = -y\log(\hat{y}) - (1-y) \log(1- \hat{y})$.

\paragraph{Weighted cross-entropy.} To facilitate learning from highly imbalanced datasets, a common adjustment to the training objective consists of reweighting terms in the cross-entropy objective:
\begin{equation*}
\end{equation*}
where hyperparameter $\omega$ determines the contribution of each class to the loss. Balanced cross-entropy is a special case of this objective, where weights are based on the prevalence of each class ($\omega$ is set as the inverse of the proportion of positive labels). Regular cross-entropy corresponds to the case where $\omega=1/2$.

\paragraph{Focal loss.} Denoting our output prediction as $\hat{y} = p_{\theta}(y = 1)$, the focal loss objective for binary classification of target $y$ is a variant on the balanced cross-entropy loss:
\begin{equation*}
    L^{focal}(y, \hat{y}) = - \omega (1-\hat{y})^{\zeta}y\log (\hat{y}) - (1-\omega) \hat{y}^{\zeta}(1-y)\log (1-\hat{y})
\end{equation*}
where $\omega_{y}$ is a balancing weight for class $y$ and $\zeta$ is the focal loss weight.

\paragraph{Multi-horizon prediction.} As highlighted in Section \ref{sec:link_mhp}, multi-horizon training can be formalized as the following objective:
    \begin{equation*}
            L^{MHP}(\mathbf{y}_{t},\hat{\mathbf{y}}_{t}) = -\frac{1}{H}\sum_{k=1}^H y^{h_k}_{t}\log(\hat{y}^{h_k}_{t}) + (1-y^{h_k}_{t})\log(1 - \hat{y}^{h_k}_{t})
    \end{equation*}
where true labels and model predictions are given by $\mathbf{y}_{t} = [y^{h_1}_{t}, \ldots, y^{h}_{t}, \ldots, y^{h_H}_{t}]$ and $\hat{\mathbf{y}}_{t} = [\hat{y}^{h_1}_{t}, \ldots, \hat{y}^{h}_{t}, \ldots, \hat{y}^{h_H}_{t}]$, for $H$ distinct horizons.

\paragraph{Label smoothing.} As introduced by \citet{DBLP:conf/cvpr/SzegedyVISW16}, label smoothing consists of substituting the original label distribution $\delta_{y = c}$ in the cross-entropy objective $L^{CE}(y,\hat{y})$ by a smoothed version $q^{LS}(c|y)$. This surrogate distribution over classes $c$ is defined as follows :
\begin{equation}
q^{LS}(c|y) = \delta_{y = c}(1-\alpha) + u(c) \alpha\\
\end{equation}
In the original approach, $u$ is uniform and $\alpha \in [0,1]$ controls the smoothing strength. By shifting the minimum of the objective function away from $\hat{y} = 1$, labels smoothing prevents the model from becoming overconfident during training. Alternative designs for $u$ have been proposed \citep{DBLP:conf/aistats/LiDB20,DBLP:conf/acl/MeisterSC20,DBLP:conf/aaai/LienenH21} but are incompatible with the binary nature of adverse event prediction. In binary tasks, labeling is defined according to the positive class such that $y \in \{0,1\}$ and $\hat{y} =  p_\theta(y = 1)$. Label smoothing therefore becomes a linear interpolation with parameter $\alpha$ such that $q^{LS} = p(y=1)$:
\begin{equation}
q^{LS} =(1-\alpha)y + \alpha(1-y)\\
\end{equation}
As suggested by \citet{DBLP:conf/icml/LukasikBMK20}, label smoothing can be used to regularize early prediction models due to the inherently noisy nature of the task. It does not, however, account for the time dependency between samples of a given stay -- highlighted in our problem formalism (Section~\ref{sec:problem_formalism}). In contrast, temporal label smoothing modulates smoothing based on time $t$ to infuse this prior knowledge into the training objective.

\subsection{DSA Objective Functions}

In this section, we detail how despite existing differences between EEP and DSA, we can train a model with a DSA objective while using it for EEP tasks at inference time. We then describe in detail the two baselines we consider from DSA: landmarking and TCSR.
\paragraph{From Survival Analysis to Early Event Prediction.} Survival analysis is a statistical framework to model the time $T$ until an event of
interest occurs. This event is considered to be terminal, thus, it is \textbf{unique} and no observation is carried after it. In survival analysis, we assume access only to an initial observation of a patient state $\X_{i,0} = [\mathbf{x}_{i,0}]$, a survival time $T_i$ and a censoring indicator $c_i$. If a patient was (right-)\textit{censored}, thus did not experience an event before the last know survival time at $T_i$, then $c_i = 1$. Otherwise, we have that $c_i = 0$, which means the patient reached a terminal state at $T_i$. Given these, we can define three probability functions:
\begin{align*}
    \text{probability mass function:} \quad  \quad  \quad f(k|\mathbf{X}) &= P(T = k | \mathbf{X}_{i,0})\\
    \text{survival function:} \quad \quad  \quad  S(k|\mathbf{X}) &= P(T> k | \mathbf{X})\\
    \text{hazard function:} \quad \quad \quad   \lambda(k|\mathbf{X}) &= P(T = k | T \geq k,\mathbf{X}) = P(T = k | T > k - 1,\mathbf{X})\\
\end{align*}
Then, if we consider only non-censored and right-censored patients, the survival likelihood can be defined as follows:
\begin{align*}
 \mathcal{L}_{surv} &= \prod_i  P(T = T_i | \mathbf{X}_{i,0})^{1-c_i}  P(T > T_i | \mathbf{X}_{i,0})^{c_i}
\end{align*}
Thus, when maximizing $\mathcal{L}_{surv}$, we aim to maximize the probability of failure at time $T_i$ if the event occurred, or the probability of survival until at least $T_i$ if the patient is censored. 
It has been shown~\citep{kalbfleisch2011statistical,craig2021survival} that MLE on $\mathcal{L}_{surv}$ is equivalent to minimizing the binary cross-entropy between hazard function estimates and labels of the form $\lambda^h_i = \mathds{1}\left[T_i = h \land c_i=0\right]$. Thus, in practice when training a model with a survival likelihood, we minimize $ \sum^N_{i=1} \sum^{T_i}_{h=1} - \lambda^h_i\log(\hat{\lambda}(h|\mathbf{X}_{0,i}))$.
As mentioned in Section~\ref{sec:pf_relatedwork}, using existing relation between $f, S$ and $\lambda$, such as $f(k|\mathbf{X}) = h(k|\mathbf{X})S(k-1|\mathbf{X})$ and  $S(k|\mathbf{X}) = 1 - \sum^k_{p=1}f(p|\mathbf{X}) = \prod^k_{p=1}(1-h(p|\mathbf{X}))$, we can recover the model's probability estimate for an event to occur within a fixed horizon $h$ as $\hat{y}^h = 1-\hat{S}(h|\mathbf{X})$.

\paragraph{Landmarking.} When multiple observations are available for a given patient, thus $\X_{i,t} = [\mathbf{x}_{i,0}, ..., \mathbf{x}_{i,t}]$, as in EEP, existing works \cite{van2007dynamic,parast2014landmark} have extended survival analysis to this dynamic context. This field is referred to as "dynamic survival analysis". As mentioned in Section~\ref{sec:pf_relatedwork}, the most prominent technique to leverage these additional observations is landmarking, where the model is fitted with new triplets of the form $(\X_{i,t}, T_i - t, c_i)$. As in regular survival analysis, when using landmarking, we minimize binary cross-entropy on the hazard function of the form   $ \sum^N_{i=1} \sum^{T_i-1}_{i=t}  \sum^{T_i- t}_{h=1} - \lambda^h_{t,i}\log(\hat{\lambda}(h|\mathbf{X}_{t,i}))$ with  $\lambda^h_{t,i} = \mathds{1}\left[T_i - t = h \land c_i=0\right]$.
As in regular survival analysis, we can recover the model's probability estimate for an event to occur within a fixed horizon $h$ from a given timepoint $t$ as $\hat{y}^h_t = 1-\hat{S}(h|\mathbf{X}_t)$, which is the probability of interest in EEP tasks.

\paragraph{Temporally consistent survival regression.} Concurrently to our work, \citet{Maystre2022} proposed TCSR, a method based on a temporally consistent dynamic sample reweighting  and label softening. Indeed, to enforce models estimate to match constraints from Equation~\ref{eq:time_consistency}, TCSR proposes to replace landmarking labels $[\lambda^1_{t,i}, \lambda^2_{t,i}, ..., \lambda^{T_i-t}_{t,i}]$ by $[\lambda^1_{t,i}, \hat{\lambda}(1|\mathbf{X}_{t-1,i}), ..., \hat{\lambda}(T_i - t - 1|\mathbf{X}_{t+1,i})]$. In addition, they also apply a reweighting according to the model estimate of the survival function, such that $ w^1_{t,i} = 1, w^2_{t,i} = 1- \hat{f}(1|\mathbf{X}_{t+1,i})$ and $w^h_{t,i} = \hat{S}(h-2|\mathbf{X}_{t+1,i}) \quad \forall h \geq 2$. 

\paragraph{Handling of non-terminal events.} Certain tasks in EEP tackles event that are terminal such as decompensation. There, the underlying assumption made in survival analysis regarding the terminality of states holds allowing to rely on a DSA approach for EEP as described above. However, in practice, most events from EEP, such as circulatory failure, are not terminal. This means that observations are carried out during and after an event. It also means other events of the same type can occur. To still use a survival analysis method for these tasks, we further split patient stays into episodes.
Using EEP notations, for a patient indexed by $i$ experiencing $v$ events at times $t_{e_1}, ..., t_{e_v}$, respectively ending at times $s_{e_1}, .. ., s_{e_v}$, we consider as distinct samples the episodes $[\mathbf{X}_{i,0},..., \mathbf{X}_{i,t_{e_1}-1}],[\mathbf{X}_{i,s_{e_1}},..., \mathbf{X}_{i,t_{e_2}-1}], ..., [\mathbf{X}_{i,s_{e_v}},..., \mathbf{X}_{i,T_i}] $. Note that this approach is consistent with EEP, where no prediction is carried out during an event.

\section{Dataset details}
\label{appendix:dataset_details}
\subsection{Task definition}
\label{appendix:task definition}

In this section, we provide more details on the definition of our early prediction tasks for {circulatory failure} from HiB \citep{yeche2021} and {decompensation} from M3B \citep{harutyunyan2019multitask}. A breakdown of event prevalence for each clinical endpoint is given in Table \ref{tab:event_prevalence}.

\begin{table}[h]
    \centering
    \caption{\textbf{Event prevalence analysis}, highlighting class imbalance. Positive timesteps are counted for 12-hour and 24-hour horizons for circulatory failure and decompensation respectively. Statistics are computed on the training set.}
    \label{tab:event_prevalence}
\begin{tabular}{lccc}
    \toprule
        \multirow{2}{*}{Task} & \multirow{2}{*}{Positive timesteps (\%)} & Patients undergoing  & Number of events \\
        & & event (\%) & per positive patient\\\midrule
        Circulatory Failure (HiRID) & 4.3 & 25.6 & 1.9 \\
        Decompensation (MIMIC) & 2.1 & 8.3 & 1.0\\
        \bottomrule
    \end{tabular}\end{table}

\textbf{Circulatory failure} is a failure of the cardiovascular system, detected in practice through elevated arterial lactate ($> 2$ mmol/l) and either low mean arterial pressure ($< 65$ mmHg) or administration of a vasopressor drug. \citet{yeche2021} defines a patient to be experiencing a circulatory failure event at a given time if those conditions are met for $2/3$ of time points in a surrounding two-hour window. Early prediction labels are then derived from these event labels as outlined in Section~\ref{sec:problem_formalism}.

\textbf{Decompensation} refers to the death of a patient. Event labels are directly extracted from the MIMIC-III \citep{johnson2016} metadata about the time of death of a patient. Early prediction labels are also extracted following Section~\ref{sec:problem_formalism}. Note that decompensation can occur outside of the ICU stay if a patient is sent to a palliative unit, for instance, which can result in patient stays with fewer than 24 positive samples.

\subsection{Pre-processing}

We describe the pre-processing steps we applied to both datasets, HiRID and MIMIC-III.

\paragraph{Imputation.}
Diverse imputation methods exist for ICU time series. For simplicity, we follow the approach of original benchmarks \citep{harutyunyan2019multitask,yeche2021} by using forward imputation when a previous measure existed. The remaining missing values are zero-imputed after scaling, corresponding to a mean imputation. 

\paragraph{Scaling.} Whereas prior work explored clipping the data to remove potential outliers \citep{tomavsev2019}, we do not adopt this approach as we found it to reduce performance on early prediction tasks. A possible explanation is that, due to the rareness of events, clipping extreme quantiles may remove parts of the signal rather than noise. Instead, we simply standard-scale data based on the training sets statistics.

\section{Implementation details}
\label{appendix:implementation_details}

\paragraph{Training details.} For all models, we set the batch size according to the available hardware capacity. Because transformers are memory-consuming, we train the {decompensation} models with a batch size of 8 stays. On the other hand, we train the GRU model for {circulatory failure} with a batch size of 64. We early stopped each model training according to their validation loss when no improvement was made after 10 epochs. 

\paragraph{Libraries.} A full list of libraries and the version we used is provided in the \texttt{environment.yml} file. The main libraries on which we build our experiments are the following: pytorch 1.11.0 \citep{NEURIPS2019_9015}, scikit-learn 0.24.1\citep{scikit-learn}, ignite 0.4.4, CUDA 10.2.89\citep{cuda}, cudNN 7.6.5\citep{chetlur2014cudnn}, gin-config 0.5.0 \citep{gin}.

\paragraph{Infrastructure.}
We follow all guidelines provided by \texttt{pytorch} documentation to ensure the reproducibility of our results. However, reproducibility across devices is not ensured. Thus we provide here the characteristics of our infrastructure. We trained all models on a single \texttt{NVIDIA RTX2080Ti} with a \texttt{Xeon E5-2630v4} core. Training took between 3 and 10 hours for a single run.

\paragraph{Uncertainty estimation.} We compute uncertainty estimates over a population of 10 training instances with different seeds. This widely-used approach has the advantage to account for the stochasticity of the training procedure, which we found to be predominant in early prediction tasks. This approach differs from other work \citep{roy2021multitask,roy2022disability,tomavsev2019,tomavsev2021} which computes uncertainty estimate by bootstrapping the test population. We found that using a pivot bootstrap estimator decreases confidence intervals by effectively increasing the population size. To be conservative with our results, we retained the former approach to compute statistics across 10 training instances. We report the 95\% confidence interval over the population means in all experiments. 

\paragraph{Architecture choices} We used the same architecture and hyperparameters reported giving the best performance on {circulatory failure} in \citet{yeche2021} and only optimized embedding regularization parameters \citep{tomavsev2019}. Exact parameters are reported in Table~\ref{tab:hp-search-gru}. For {decompensation}, as we found a transformer architecture to perform better than originally proposed models \citep{harutyunyan2019multitask}, we carried out our own random search on validation AUPRC performance. The exact parameters for this task are reported in Table~\ref{tab:hp-search-decomp}.

\begin{table}[tbh!]
    \centering
\caption{\textbf{Hyperparameter search range} for {circulatory failure} with GRU \citep{DBLP:journals/corr/ChungGCB14} backbone. In \textbf{bold} are parameters selected by random search.}
\begin{tabular}{lc}
\toprule
Hyperparameter & Values\\
\midrule
\midrule
Learning Rate & (1e-5, 3e-5, 1e-4, \textbf{3e-4}) \\
\midrule
Drop-out & (\textbf{0.0}, 0.1, 0.2, 0.3, 0.4) \\
\midrule
Depth &   (1, \textbf{2}, 3) \\
\midrule
Hidden Dimension & (32, 64, 128, \textbf{256}) \\
\midrule
L1 Regularization &  (1e-2, 1e-1, 1, \textbf{10}, 100)\\
\bottomrule
\end{tabular}
\label{tab:hp-search-gru}
\end{table}

\begin{table}[tbh!]
    \centering
\caption{\textbf{Hyperparameter search range} for {decompensation} with Transformer \citep{DBLP:conf/nips/VaswaniSPUJGKP17} backbone. In \textbf{bold} are parameters selected by random search.}
\begin{tabular}{lc}
\toprule
Hyperparameter & Values\\
\midrule
\midrule
Learning Rate & (1e-5, 3e-5, \textbf{1e-4}, 3e-4) \\
\midrule
Drop-out & (0.0, 0.1, 0.2, \textbf{0.3}, 0.4) \\
\midrule
Attention Drop-out &   (0.0, \textbf{0.1}, 0.2, 0.3, 0.4) \\
\midrule
Depth &   (1, \textbf{2}, 3) \\
\midrule
Heads &  (\textbf{1}, 2, 4) \\
\midrule
Hidden Dimension &  (32, \textbf{64}, 128, 256) \\
\midrule
L1 Regularization &  (1e-2, \textbf{1e-1}, 1, 10)\\
\bottomrule
\end{tabular}
\label{tab:hp-search-decomp}
\end{table}

\subsection{Baseline implementation}

\paragraph{Balanced cross-entropy.}
In the binary setting, the only hyperparameter of balanced cross-entropy is the relative contribution of the minority class to the loss, $\omega$. As discussed in Section~\ref{sec:Ablations}, no value of $\omega$ was found to improve validation performance over the non-balanced case $\omega=1$.

\paragraph{Focal loss.}
A grid search over focal loss hyperparameters was also carried out.
Similarly to balanced cross-entropy, on all tasks, no values of focal loss weight $\zeta$ or balancing weight $\omega$ were found to outperform regular cross-entropy corresponding to $\zeta = 0 $ and $\omega=1$.

\paragraph{Multi-horizon prediction.} Following \citet{tomavsev2019}, we consider $H$ horizons on both side of the true horizon $h$ between $0$ and $2h$. As we didn't find $H \longrightarrow +\infty$, to increase performance, we selected $H = 11$ (including true horizon $h$) compared to $H = 8$ in \citet{tomavsev2019}, which we found to perform slightly worse. This means we made a prediction every $2$ hours for circulatory failure and every $4$ hours for {decompensation}. 

\paragraph{Label smoothing.} Label smoothing \citep{DBLP:conf/cvpr/SzegedyVISW16}, as defined in Section~\ref{sec:TLS}, is normally used in multi-class setting. We still compared our method to it for two reasons. First, to explore if it can help when dealing with a noisy signal as we claim is the case for early event detection. Second, to ablate the impact of adding a temporal dependency to the method. Again, we select the hyperparameter $\alpha$ through a grid search. Interestingly, we found label smoothing to slightly improve performance over the validation set for all tasks as opposed to the results reported for the test set in Table~\ref{tab:perf_results}. We found $\alpha = 0.05$ to perform best for both {circulatory failure} and {decompensation}.

\paragraph{Landmarking.} For all tasks, landmarking was trained with the same architecture and parameters with the exception that our model return hazard estimates. In theory, we should make predictions until $h_{\max} = \max_{i}( T_i)$ corresponding to 2016 and 2805, for respectively circulatory failure and decompensation. Due to computing limation, as is common in practice, we truncated this horizon to 1000 for circulatory failure.

\paragraph{TCSR.}
As for landmarking, we considered $h_{\max}$ to be 1000 and 2805 for respectively circulatory failure and decompensation. In practice, we found that the dynamic nature of the label and weight assignment lead to great instability. To be able to train correctly models with this objective, we had to reduce learning rates to 5e-5 and 3e-5. More importantly, for circulatory failure, we used stop-gradient operation for predictions such that $\frac{d\mathcal{L}_{i,t}}{d\hat{\lambda}^h_{i,t+1}} = 0$. A similar approach for decompensation resulted in worse results, thus we did not use it for this task.

\subsection{TLS implementation}

\begin{figure}[hbtp]
\begin{lstlisting}
def get_smoothed_labels(event_label_patient, smoothing_fn, h_true, h_min,
                        h_max, **kwargs):
                        
    # Find when event label changes
    diffs = np.concatenate([np.zeros(1), 
                event_label_patient[1:] - event_label_patient[:-1]], axis=-1)
    pos_event_change = np.where((diffs == 1) & (event_label_patient == 1))[0]
    
    # Handle patients with no events
    if len(pos_event_change) == 0: 
        pos_event_change = np.array([np.inf])

    # Compute distance to closest event for each time point
    time_array = np.arange(len(event_label_patient))
    dist_all_event = pos_event_change.reshape(-1, 1) - time_array
    dist_to_closest = np.where(dist_all_event > 0,
                                  dist_all_event, np.inf).min(axis=0)

    return smoothing_fn(dist_to_closest, h_true=h_true, h_min=h_min, h_max=h_max,
                                                                    **kwargs)
\end{lstlisting}
\caption{\textbf{Temporal label smoothing algorithm.} Python-style code to obtain smooth early prediction labels from event labels.}
\label{fig:code_snippet}
\end{figure}
\begin{comment}
\begin{lstlisting}
        def q_exp(dt, h_true, h_min, h_max, delta_h, gamma):
    """Returns q(t) for alpha_exp.
    
    dt: (int) distance to next event in steps.
    h_true: (int) true horizon of prediction in steps.
    h_min: (int) minimum horizon to apply smoothing in steps.
    h_max: (int) maximum horizon to apply smoothing in steps.
    delta_h: (int) number of steps per hour.

    """
    if dt <= h_min:
        return 1
    elif dt > h_max:
        return 0
    else:
        h_min_scaled = h_min / delta_h
        h_max_scaled = h_max / delta_h
        dt_scaled = dt / delta_h

        d = -(1 / gamma) * np.log(np.exp(-gamma * (h_min_scaled)) - np.exp(-gamma * (h_max_scaled)))
        A = -np.exp(-gamma * (h_max_scaled - d))
        return np.exp(-gamma * (dt_scaled - d)) + A
\end{lstlisting}
\end{comment}

TLS depends on two components, the temporal range over which we smooth labels, defined by $h_{min}$ and $h_{max}$, and the smoothing function $q(t)$. Concerning the temporal range, for a fair comparison, we fix it to match MHP, thus for all experiments, we set $h_{min} = 0$ and $h_{max} = 2h$. For the smoothing function, we perform a grid search over the type of function discussed in Appendix~\ref{appendix:temporal_smoothing_fn} and the smoothing strength parameter $\gamma$. For all experiments, we found $q^{exp}$ to outperform other considered functions. Given validation performance, we used $\gamma = 0.2$ for {circulatory failure} and $\gamma = 0.05$ for {decompensation}.

As discussed in Section~\ref{sec:TLS}, contrary to MHP, TLS does not require any change to the architecture leading to a computational overhead. The smoothing of the labels can be easily integrated into the data loader, as shown in Figure~\ref{fig:code_snippet}. 

\begin{figure}[h]
\centering
\begin{subfigure}[b]{0.49\textwidth}
 \centering
  \includegraphics[width=\linewidth]{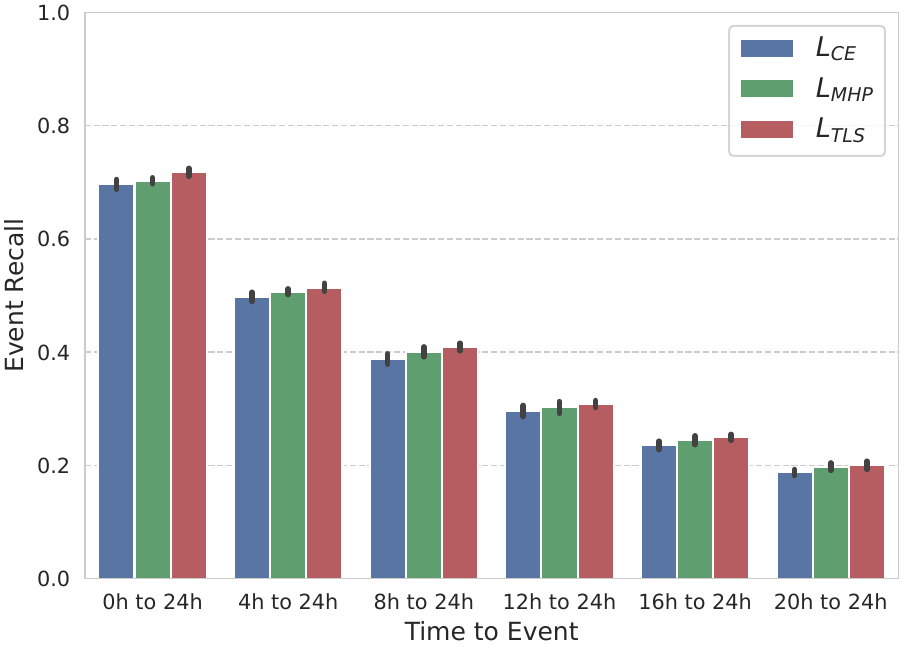}
  \caption{\textit{Event recall} at 50\% timestep-level precision.} \label{fig:event_decomp}
\end{subfigure}
\begin{subfigure}[b]{0.49\textwidth}
  \centering
  \includegraphics[width=\linewidth]{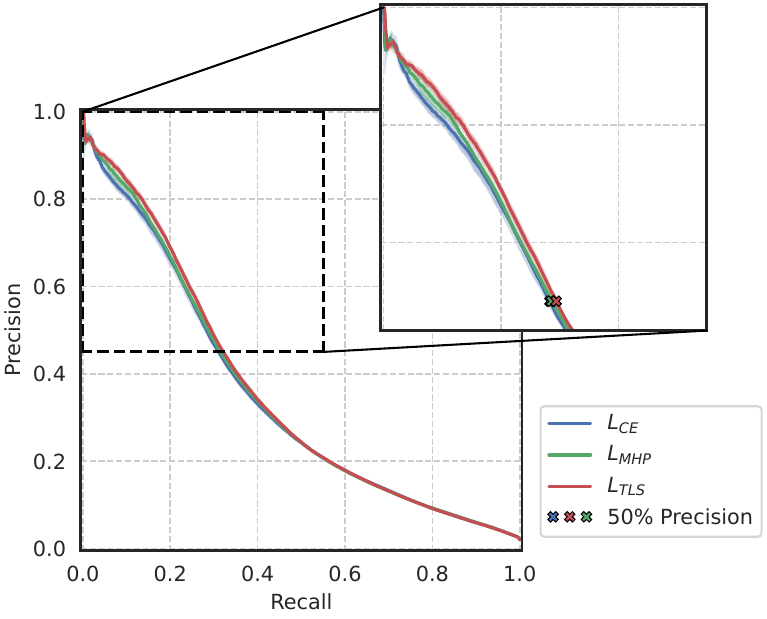}
  \caption{\textit{Precision-recall curve.}}
  \label{fig:PR_curve_decomp}
\end{subfigure}
\caption{\textbf{Clinically relevant performance} on decompensation. Inset in (b) shows the clinically-applicable region with precision greater than 50\%.}
\label{fig:clinical_performance_decomp}
\end{figure}

\section{Additional experiments and ablation studies}
\label{appendix:add_exp}

This section provides additional results and experiments to complete our findings from the main manuscript. Unless otherwise stated, mean results are shown with a 95\% confidence interval on the mean shaded or in error bars.

\subsection{Performance analysis for decompensation prediction}
\label{appendix:decomp_results}
Event-level performance for decompensation prediction is given in Figure \ref{fig:event_decomp}. Results are similar to those on circulatory failure discussed in Section \ref{sec:perf-results}: temporal label smoothing improves recall of adverse event episodes over cross-entropy and MHP. Note that the improvements observed over the baselines in terms of event-recall between 0 and $h$ are smaller than for circulatory failure, but are statistically significant as shown in Table\ref{tab:perf_results}

\begin{figure}[h]
\begin{subfigure}[b]{0.47\textwidth}
  \centering
  \includegraphics[width=\linewidth]{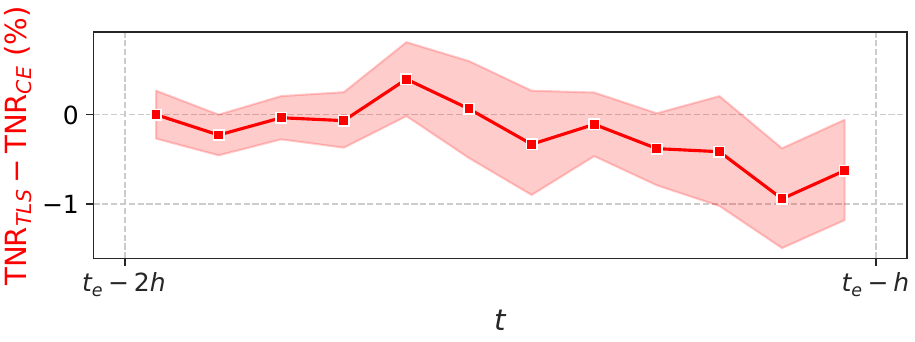}\vspace{-0.5em}\caption{True negative rate (TNR).}
  \label{fig:delta_tnr_decomp}
\end{subfigure} \hfill
\begin{subfigure}[b]{0.47\textwidth}
 \centering
  \includegraphics[width=\linewidth]{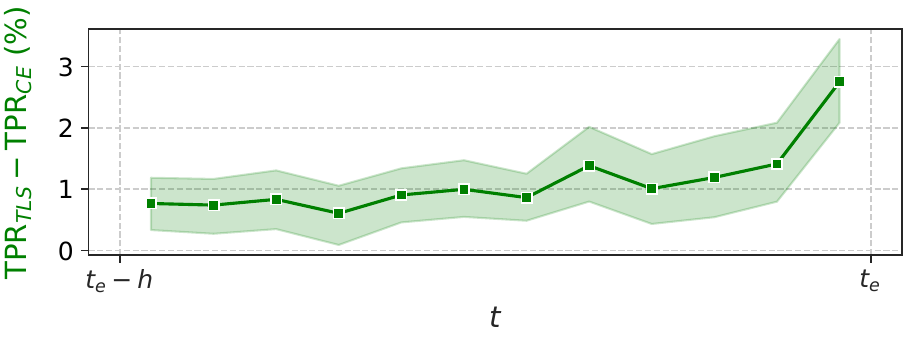}\vspace{-0.5em}\caption{True positive rate (TPR).}
  \label{fig:delta_tpr_decomp}
\end{subfigure}
\caption{\textbf{Performance improvement over time} for TLS over traditional cross-entropy on decompensation prediction. Timestep-level metrics computed for precision of $0.5$ over two-hour bins.}
\vspace{-1em}
\label{fig:plot_delta_decomp}
\end{figure}

The precision-recall curve obtained for timestep-level event prediction on this task is also given in Figure \ref{fig:PR_curve_decomp}. As for circulatory failure prediction, recall gains are concentrated in regions of low false-alarm rates (>50\% precision) which are most clinically relevant. 

Likewise, whereas recall near the label boundary $t_e -h$ is slightly negatively affected by temporal label smoothing in Figure \ref{fig:plot_delta_decomp}, true positive rates are significantly improved leading up to the event time $t_e$. This mirrors the temporal smoothing pattern which favors higher model confidence away from the label boundary. As discussed in Section \ref{sec:Ablations}, this is aligned with clinical priorities in terms of model performance, as it ensures imminent events are better predicted.

\subsection{Sub-group analysis}
Populations in the intensive care unit are often heterogeneous. This has motivated recent works to focus on the fairness of deep learning across these sub-populations. In this analysis, we ensure that temporal label smoothing does not negatively affect performance in specific subgroups, compared to the objectives commonly used in the literature \citep{tomavsev2019, hyland2020, Lauritsen2020}. To achieve this, we measured event prediction performance across genders and age groups (below 50, between 50 and 70, and over 70 years old). As shown in Table~\ref{tab:sg-circ}, TLS matches or outperforms baseline performance across all studied subgroups, suggesting that the overall population-wide improvements are not achieved by disproportionally favouring specific cohorts. While some algorithmic bias can be observed across all methods, for instance in poorer decompensation performance amongst female patients, TLS does not appear to be amplifying this issue. In further work, we look forward to extending this analysis to more specific subgroups and studying the fairness of early event prediction methods for clinical applications.

\begin{table}[h]
{
 \centering
    \caption{\textbf{Sub-group performance analysis.} We color improvement above the 95\% confidence interval in {\color{Green} green}.} \label{tab:sg-circ}
\resizebox{\textwidth}{!}{\begin{tabular}{lcccccccccc}
\toprule
 Circulatory Failure & \multicolumn{2}{c}{Age $\leq 50$} & \multicolumn{2}{c}{$50 <$ Age $\leq 70$} & \multicolumn{2}{c}{Age $> 70$ } & \multicolumn{2}{c}{Female} & \multicolumn{2}{c}{Male}\\
 \cmidrule(lr){2-3} \cmidrule(lr){4-5}\cmidrule(lr){6-7}\cmidrule(lr){8-9}\cmidrule(lr){10-11}
Method &         AUPRC &   Recall &         AUPRC &   Recall &         AUPRC &   Recall & AUPRC &   Recall & AUPRC &   Recall\\
\midrule
CE     &             40.4 $\pm$ 0.5 &             29.4 $\pm$ 0.6 &             38.8 $\pm$ 0.6 &             29.6 $\pm$ 1.1 &             39.2 $\pm$ 0.3 &             29.0 $\pm$ 1.0 &             39.3 $\pm$ 0.6 &             30.0 $\pm$ 0.7 &             39.1 $\pm$ 0.4 &             29.0 $\pm$ 1.0 \\
\textbf{TLS}     & 40.4 $\pm$ 0.5 &  $\mathbf{32.7}$ $\pm$ 1.0 &  $\mathbf{41.1}$ $\pm$ 0.4 &  $\mathbf{32.6}$ $\pm$ 0.7 &  $\mathbf{40.0}$ $\pm$ 0.3 &  $\mathbf{31.7}$ $\pm$ 0.7 &  $\mathbf{41.2}$ $\pm$ 0.3 &  $\mathbf{32.8}$ $\pm$ 0.6 &  $\mathbf{40.4}$ $\pm$ 0.3 &  $\mathbf{32.0}$ $\pm$ 0.8 \\
$\Delta$(TLS-CE)    &  {0.0}&  {\color{Green}+ 3.3} &  {\color{Green}+ 2.3} &  {\color{Green}+ 3.0} &  {\color{Green}+ 0.9} &  {\color{Green}+ 2.7} &  {\color{Green}+ 1.8} &  {\color{Green}+ 2.9} &  {\color{Green}+ 1.3} &  {\color{Green}+ 3.0} \\
\bottomrule
\end{tabular}}
    \centering
\resizebox{\textwidth}{!}{\begin{tabular}{lcccccccccc}

\toprule
 Decompensation & \multicolumn{2}{c}{Age $\leq 50$} & \multicolumn{2}{c}{$50 <$ Age $\leq 70$} & \multicolumn{2}{c}{Age $> 70$ } & \multicolumn{2}{c}{Female} & \multicolumn{2}{c}{Male}\\
 \cmidrule(lr){2-3} \cmidrule(lr){4-5}\cmidrule(lr){6-7}\cmidrule(lr){8-9}\cmidrule(lr){10-11}
Method &         AUPRC &   Recall &         AUPRC &   Recall &         AUPRC &   Recall & AUPRC &   Recall & AUPRC &   Recall\\
\midrule
CE     &              29.2 $\pm$ 0.8 &             25.3 $\pm$ 1.2 &             34.9 $\pm$ 0.9 &             27.4 $\pm$ 0.6 &             35.8 $\pm$ 0.2 &             29.4 $\pm$ 0.6 &             30.9 $\pm$ 0.4 &             24.8 $\pm$ 0.6 &             38.3 $\pm$ 0.6 &             31.4 $\pm$ 0.5 \\
\textbf{TLS}     &  $\mathbf{30.5}$ $\pm$ 0.5 &             26.2 $\pm$ 1.1 &  $\mathbf{36.7}$ $\pm$ 0.5 &  $\mathbf{29.1}$ $\pm$ 0.5 &  $\mathbf{36.3}$ $\pm$ 0.3 &  $\mathbf{30.3}$ $\pm$ 0.4 &  $\mathbf{31.6}$ $\pm$ 0.3 &  $\mathbf{25.7}$ $\pm$ 0.5 &  $\mathbf{39.6}$ $\pm$ 0.5 &  $\mathbf{32.8}$ $\pm$ 0.6 \\
$\Delta$(TLS-CE)    &  {\color{Green}+ 1.3} &  {\color{Green}+ 1.0} &  {\color{Green}+ 1.8} &  {\color{Green}+ 1.7} &  {\color{Green}+ 0.5} &  {\color{Green}+ 0.9} &  {\color{Green}+ 0.7} &  {\color{Green}+ 0.9} &  {\color{Green}+ 1.3} &  {\color{Green}+ 1.4} \\
\bottomrule
\end{tabular}}}
\end{table}

\subsection{Loss reweighting methods}
\label{appendix:loss_reweight}

\begin{figure}[h]
\centering
\begin{subfigure}[b]{0.32\textwidth}
  \centering
\includegraphics[width=\linewidth]{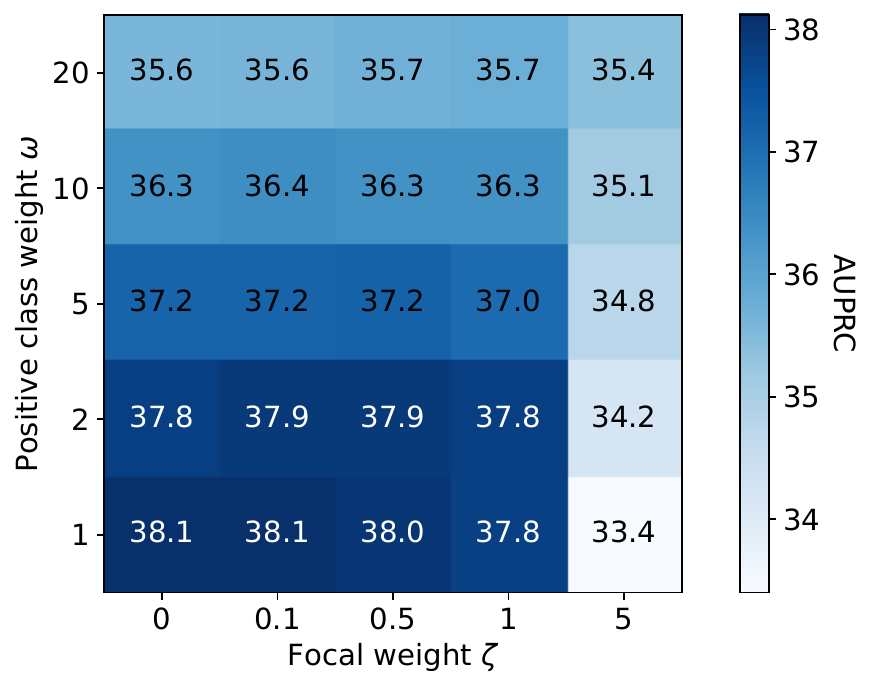} 
  \caption{\textit{Reduction in AUPRC (validation).}}
\label{fig:grid_focal_decomp}
\end{subfigure} \hspace{4em}
\begin{subfigure}[b]{0.32\textwidth}
  \centering
  \includegraphics[width=\linewidth]{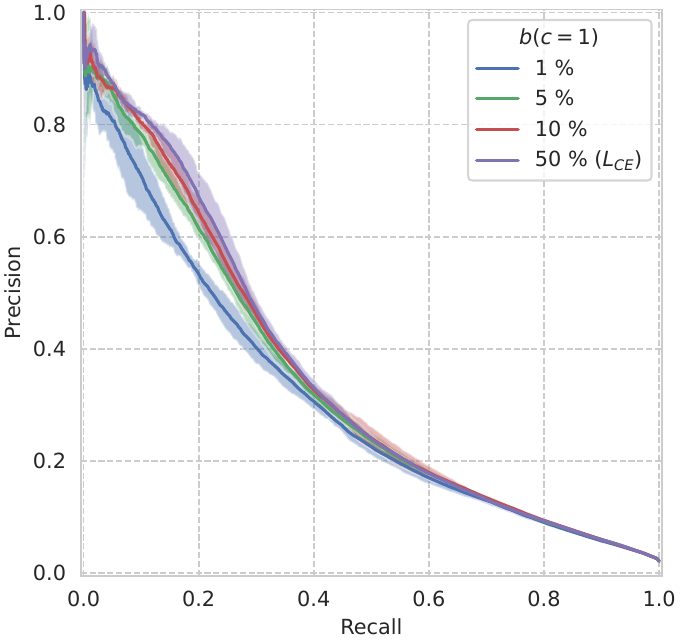}
  \caption{\textit{Reduction in precision.}}
  \label{fig:rw_PR_curve_decomp}
\end{subfigure}
\caption{\textbf{Performance loss with class reweighting methods}, on decompensation prediction. (a) Balanced cross-entropy corresponds to $\zeta=0$, focal loss to $\zeta \geq 0$. (b) Loss reweighting does not improve AUPRC because it significantly reduces precision. Balance weights correspond to $b(c)$. Similar results for circulatory failure prediction.}
\label{fig:additional_PR_reweight}
\end{figure}

Hyperparameter grid search results on decompensation prediction for different loss reweighting methods are shown in Figure~\ref{fig:grid_focal_decomp}. Weighted cross-entropy and focal loss were also found to negatively affect performance in comparison to traditional cross-entropy. Likely explanations for these results are provided in Section \ref{sec:Ablations}: focal loss focuses training on noisily labeled samples, and weighted cross-entropy largely reduces precision.

We validate the latter hypothesis by visualizing precision-recall curves of models trained with this objective in Figure \ref{fig:additional_PR_reweight}. With a relative weight for the positive class $\omega > 1$, weighted cross-entropy encourages a greater number of true positives to improve recall. Doing so also increases the of false positives, impairing precision. In Figure \ref{fig:additional_PR_reweight}, as the starting precision of all cross-entropy models is poor, no discernible improvements in the recall can be observed as class weights are increased, whereas precision is markedly reduced in low-recall regions. This explains the overall reduction in AUPRC with this method.

\newpage
\section{Alternative early prediction tasks}\label{appendix:resp}

As a third task to benchmark our method, we studied early prediction of respiratory failure, defined in \citet{yeche2021}. Unfortunately, this task has vague labels which result in all methods performing close to random. For transparency, we first provide results on this task and motivate our belief that this label ambiguity is caused by a very noisy estimate of a certain clinical variable (FIO$_2$). See Section \ref{sec:label-issues} for details. 

As a related, alternative dataset, we define a related sub-task that does not rely on FIO$_2$: prediction of the onset of mechanical ventilation. For this task, we show that: (1) models do perform much better than random, which confirms our hypothesis on respiratory failure labeling, and (2) TLS improves again significantly over EEP baselines, with similar results to in Section~\ref{sec:perf-results}.

 \paragraph{Implementation details.} For respiratory failure prediction, we used the transformer architecture and hyperparameters for {respiratory failure} reported in \citet{yeche2021}. For ventilation onset, we used a GRU model and selected hyperparameters based on a grid search over the validation AUPRC. This resulted in a 2-layer GRU with a hidden space dimensionality of 128 and no dropout. In both cases, we chose 10.0 as the $l_1$ regularization strength for the embedding module and used a batch size of 8 stays. For label smoothing, we found $\alpha = 0.1$ to give the best validation performance. We used $\gamma=0.05$ (respiratory failure) and $\gamma=0.1$ (ventilation onset) for temporal label smoothing with exponential parametrization.
 
\subsection{Labeling issues for respiratory failure}
\label{sec:label-issues}

Respiratory failure is defined as a P/F ratio (arterial pO$_2$ over FIO$_2$) below $300$ mmHg \citep{yeche2021}. This includes mild failure events, which results in high event prevalence in the HIRID dataset \citep{hyland2020}: 38.6\% of timepoints have a positive label, and 83\% of patients undergo at least one event, with on average 1.8 events per positive patient. Despite this high prevalence, all EEP methods have a performance close to $60\%$ AUPRC, as shown in Table~\ref{tab:perf_results_resp_vent} and as in~\citet{yeche2021}. This corresponds to an enrichment factor (ratio of AUPRC of predictor vs.\ random classifier) with respect to a random classifier ($\approx 40\%$) of $~1.5$ for this task, compared to factors of $~10$ and $~15$ for circulatory failure and decompensation, respectively. Such a low performance suggests an inherent issue with labeling. Our hypothesis is that the estimation of FIO$_2$ is highly error-prone, which challenges the quality of respiratory failure labels and causes the low performance of all machine learning models considered. For completeness, we nevertheless show the results for respiratory failure (in addition to ventilation onset in this section and circulatory failure as well as decompensation in the main part).

\subsection{Ablation study: onset of mechanical ventilation}
 To verify the above hypothesis, we define a similar task independent of FIO$_2$ estimates and verify we can recover a better baseline performance. We focus on predicting whether a patient will be mechanically ventilated within the next 12 hours. Ventilation is a good proxy for severe respiratory distress but is not labeled based on a P/F ratio estimate. With a 5.6\% timestep-level prevalence, baseline performance at $34\%$ AUPRC in Table \ref{tab:perf_results_resp_vent} is roughly 6.2 times better than a random classifier. This confirms that poor FIO$_2$ estimation underlies poor performance on respiratory failure prediction across all methods.

\begin{comment}
\begin{table}[tbh!]
    \centering
\caption{\textbf{Hyperparameter search range} for {respiratory failure} with Transformer \citep{DBLP:conf/nips/VaswaniSPUJGKP17} backbone. In \textbf{bold} are parameters selected by random search.}
\begin{tabular}{lc}
\toprule
Hyperparameter & Values\\
\midrule
\midrule
Learning Rate & (1e-5, 3e-5, \textbf{1e-4}, 3e-4) \\
\midrule
Drop-out & (0.0, 0.1, 0.2, \textbf{0.3}, 0.4) \\
\midrule
Attention Drop-out &   (\textbf{0.0}, 0.1, 0.2, 0.3, 0.4) \\
\midrule
Depth &   (1, \textbf{2}, 3) \\
\midrule
Heads &  (\textbf{1}, 2, 4) \\
\midrule
Hidden Dimension &  (32, \textbf{64}, 128, 256) \\
\midrule
L1 Regularization &  (1e-2, 1e-1, 1, \textbf{10}, 100)\\
\bottomrule
\end{tabular}
\label{tab:hp-search-resp}
\end{table}
\end{comment}

\begin{table*}[h] \centering
    \caption{\textbf{Performance of different training objectives for early prediction of respiratory failure and ventilation onset.} Recall is reported at a 50\% timestep-level precision. {In \textbf{bold}, we highlight best-performing methods with statistically significant $p$-values ($<0.05$) under paired Student's t-tests~\citep{student1908probable}} compared with the next-best method marked in italic.}
    \label{tab:perf_results_resp_vent}
\resizebox{\textwidth}{!}{
\begin{tabular}{lcccccc}
\toprule
 Task & \multicolumn{3}{c}{Respiratory Failure (HiRID)} & \multicolumn{3}{c}{Ventilation Onset (HiRID)} \\
 \cmidrule(lr){2-4} \cmidrule(lr){5-7}
Training objective &         AUPRC & Timestep  Recall & Event Recall & AUPRC & Timestep  Recall & Event Recall  \\
\midrule
Cross-entropy   \citep{Lauritsen2020, hyland2020}  &  {60.5} $\pm$ 0.2 &  {77.3} $\pm$ 0.5 & 94.9 $\pm$ 0.2 & 34.1 $\pm$ 0.4 &           {\it 23.0} $\pm$ 1.1& 64.2 $\pm$ 1.8 \\
Multi-horizon \citep{tomavsev2019,jarrett2019dynamic} &             {60.3} $\pm$ 0.1 &             ${76.6}$ $\pm$ 0.5  & {\it 95.0} $\pm$ 0.1 &  {\it 34.4} $\pm$ 0.5 &           {\it 23.0} $\pm$ 0.6 & {\it 64.3} $\pm$ 0.9 \\
\textbf{Temporal Label Smoothing}    &   {\it 60.4} $\pm$ 0.2 &  {\it 77.0} $\pm$ 0.3  & \textbf{95.3} $\pm$ 0.1   & 34.7 $\pm$ 0.4 &          \textbf{ 24.2} $\pm$ 0.7 &	\textbf{67.8} $\pm$ 0.9\\
\midrule
$p$-value & {0.15} & {0.14} &  \textbf{0.04} & 0.25 & \textbf{0.008} & 
\textbf{<0.001} \\
\midrule
 Enrichment Factor & \multicolumn{3}{c}{\textbf{1.5}} & \multicolumn{3}{c}{\textbf{6.2}}\\
\bottomrule
\end{tabular}}
\end{table*}

\subsection{Temporal label smoothing performance for onset of mechanical ventilation}

\begin{figure}[h]
\centering
\begin{subfigure}[b]{0.49\textwidth}
 \centering
  \includegraphics[width=\linewidth]{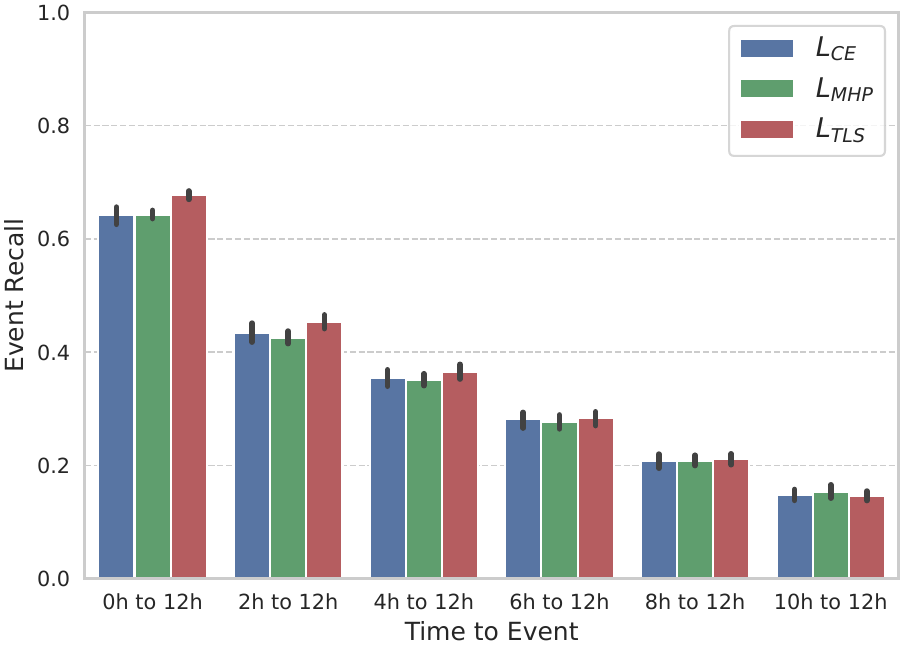}
  \caption{\textit{Event recall} at 50\% timestep-level precision.} \label{fig:based_vent}
\end{subfigure}
\begin{subfigure}[b]{0.49\textwidth}
  \centering
  \includegraphics[width=\linewidth]{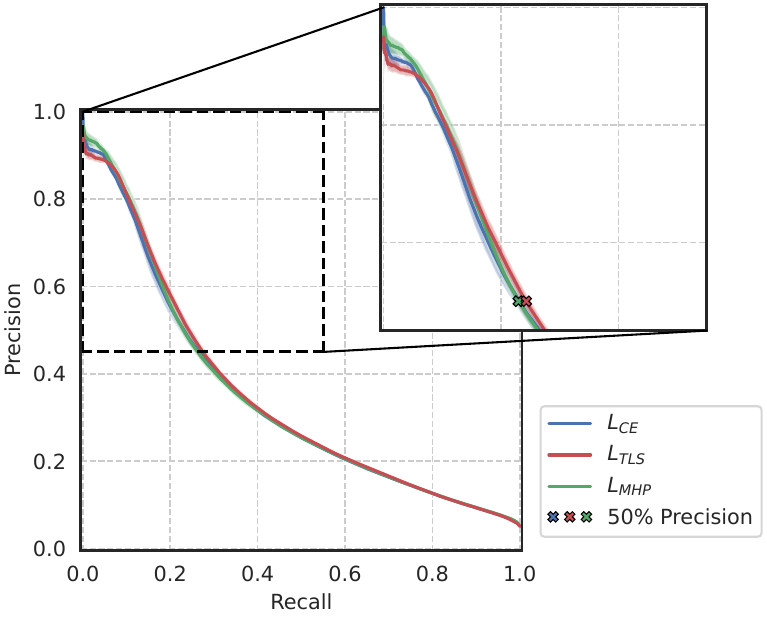}
  \caption{\textit{Precision-recall curve.} Inset shows the clinically-applicable region with precision greater than $0.5$}
  \label{fig:PR_curve_vent}
\end{subfigure}
\caption{\textbf{Clinically relevant performance} on ventilation onset.}
\label{fig:clinical_performance_vent}
\end{figure}

\begin{figure}[h]
\begin{subfigure}[b]{0.47\textwidth}
  \centering
  \includegraphics[width=\linewidth]{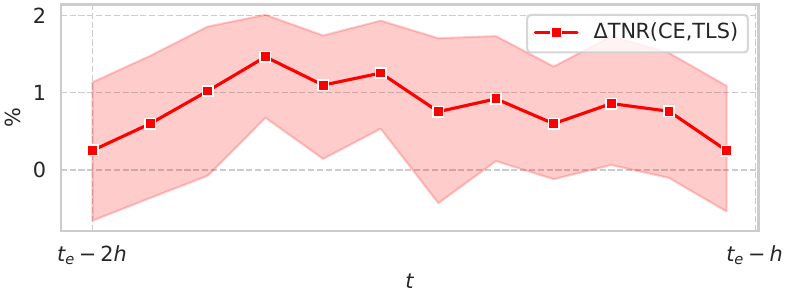}\vspace{-0.5em}\caption{True negative rate (TNR)}\vspace{-0.25em}
  \label{fig:delta_tnr_vent}
\end{subfigure} \hfill
\begin{subfigure}[b]{0.47\textwidth}
 \centering
  \includegraphics[width=\linewidth]{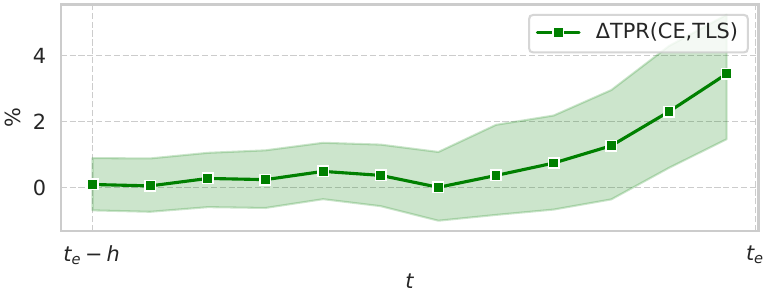}\vspace{-0.5em}\caption{True positive rate (TPR)}\vspace{-0.25em}
  \label{fig:delta_tpr_vent}
\end{subfigure}
\caption{\textbf{Performance improvement over time} for TLS over traditional cross-entropy on onset ventilation prediction. Timestep-level metrics computed for precision of $0.5$ over two-hour bins.}
\vspace{-1em}
\label{fig:plot_delta_vent}
\end{figure}

In this final section, we verify the benefits of TLS in predicting the onset of mechanical ventilation -- a feasible task relative to the respiratory system. In Table~\ref{tab:perf_results_resp_vent}  and Figure~\ref{fig:based_vent}, we find that TLS again improves performance in both timestep and event recall over multi-horizon prediction, and performs on par in terms of AUPRC. This is likely due to its lower performance at very low recall in Figure~\ref{fig:PR_curve_vent}. Finally, TLS again improves the true negative and positive rates away from the label boundary $t_e-h$ in Figure~\ref{fig:plot_delta_vent}, which corresponds to more clinically relevant regions. All conclusions agree with our analysis on other tasks in Section \ref{sec:Ablations}.

\end{document}